\documentclass[runningheads]{llncs}

\usepackage{eccv}

\usepackage{eccvabbrv}

\usepackage{graphicx}
\usepackage{booktabs}

\usepackage{booktabs}
\usepackage{tabularray}
\usepackage{multicol}
\usepackage{multirow}
\usepackage{caption}
\usepackage{graphicx}
\usepackage{listings}
\usepackage{makecell}
\usepackage{chapterbib}
\usepackage{bbding}
\usepackage{enumitem}
\UseTblrLibrary{booktabs}
\usepackage{wasysym}             
\usepackage[table]{xcolor}
\usepackage{nicematrix}
\usepackage{adjustbox}
\usepackage{xspace}
\usepackage{cuted}
\usepackage[ruled,vlined]{algorithm2e}

\setlist[itemize, 1]{label =\raisebox{-0.1\height}{\scalebox{1.3}{\textbullet}}}
\setlist[itemize]{noitemsep, topsep=-0.05cm, leftmargin=4mm}

\newcommand{\Sref}[1]{Sec.~\ref{#1}}
\newcommand{\Eref}[1]{Eq.~(\ref{#1})}
\newcommand{\Fref}[1]{Fig.~\ref{#1}}
\newcommand{\Tref}[1]{Tab.~\ref{#1}}
\newcommand{\R}[0]{\mathbb{R}}
\newcommand{\Lo}[0]{\mathbb{L}}

\newcommand{\fst}[1]{\textbf{#1}}

\newcommand{\inc}[1]{{\color{gain}#1}}
\newcommand{\dec}[1]{{\color{lost}#1}}
\newcommand{\pdif}[1]{{\color{gain}\scriptsize{(+#1)}}}

\newcommand\mypara[1]{\vspace{1mm}\noindent\textbf{#1}.\!}

\definecolor{mypink}{rgb}{0.95, 0.95, 1.0}
\definecolor{myorange}{rgb}{0.98, 0.98, 0.98}

\definecolor{cvprblue}{rgb}{0.21,0.49,0.74}

\definecolor{gain}{HTML}{34a853}
\definecolor{lost}{HTML}{ea4335}
\definecolor{theme_color}{rgb}{0.75, 0.22, 0.46}
\definecolor{gt}{HTML}{cc6092}

\providecommand\rightarrowRHD{\relbar\joinrel\mathrel\RHD}

\newcommand{\uparrowRHD}  {\rotatebox[origin=c]{90}{$\rightarrowRHD$}}
\newcommand{\downarrowRHD}{\rotatebox[origin=c]{270}{$\rightarrowRHD$}}

\newcommand{\uparrowRHDSmall}  {\raisebox{0.05\normalbaselineskip}{\scalebox{0.7}{\uparrowRHD}}}   
\newcommand{\downarrowRHDSmall}{\raisebox{0.07\normalbaselineskip}{\scalebox{0.7}{\downarrowRHD}}} 

\newcommand{\methodName}{ARGENT\xspace}
\newcommand{\vlm}{VLM\xspace}
\newcommand{\vlms}{VLMs\xspace}

\newcommand{\hycoclip}{HyCoCLIP\xspace}
\newcommand{\clip}{CLIP\xspace}
\newcommand{\meru}{MERU\xspace}

\newcommand{\grit}{GRIT\xspace}
\newcommand{\coco}{COCO\xspace}
\newcommand{\sota}{SOTA\xspace}

\usepackage[accsupp]{axessibility}  

\usepackage[pagebackref,breaklinks,colorlinks,citecolor=eccvblue]{hyperref}

\usepackage{orcidlink}

\begin{document}


\title{\methodName: Adaptive Hierarchical Image-Text Representations}

\titlerunning{ARGENT}

\author{Chuong Huynh\inst{1}\textsuperscript{\dag} \and
Hossein Souri\inst{2} \and
Abhinav Kumar\inst{2} \and 
Vitali Petsiuk\inst{2} \and \\
Deen Dayal Mohan\inst{2} \and 
Suren Kumar\inst{2}\textsuperscript{\dag}
}

\authorrunning{C.~Huynh et al.}

\institute{University of Maryland, College Park \and
Samsung Research America, AI Center -- Mountain View\\
Project Page: \url{https://hmchuong.github.io/argent}}

\maketitle

\begin{figure}[t]
    \centering
    \begin{minipage}[b]{0.63\linewidth}
        \centering
        \includegraphics[width=\linewidth]{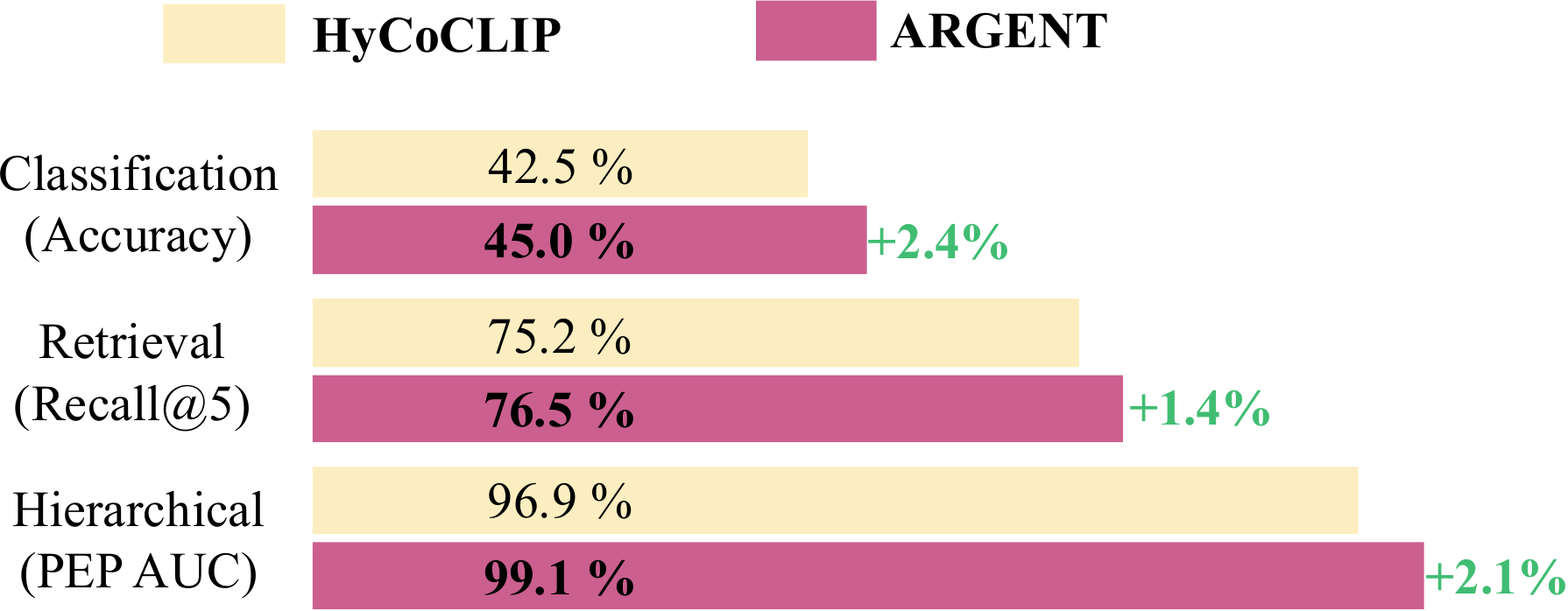}
        \vspace{-0.2cm}
        \vfill
        {\small (a) \methodName outperforms \sota \hycoclip.}
    \end{minipage}
    \hfill 
    \begin{minipage}[b]{0.33\linewidth}
        \centering
        \includegraphics[width=\linewidth]{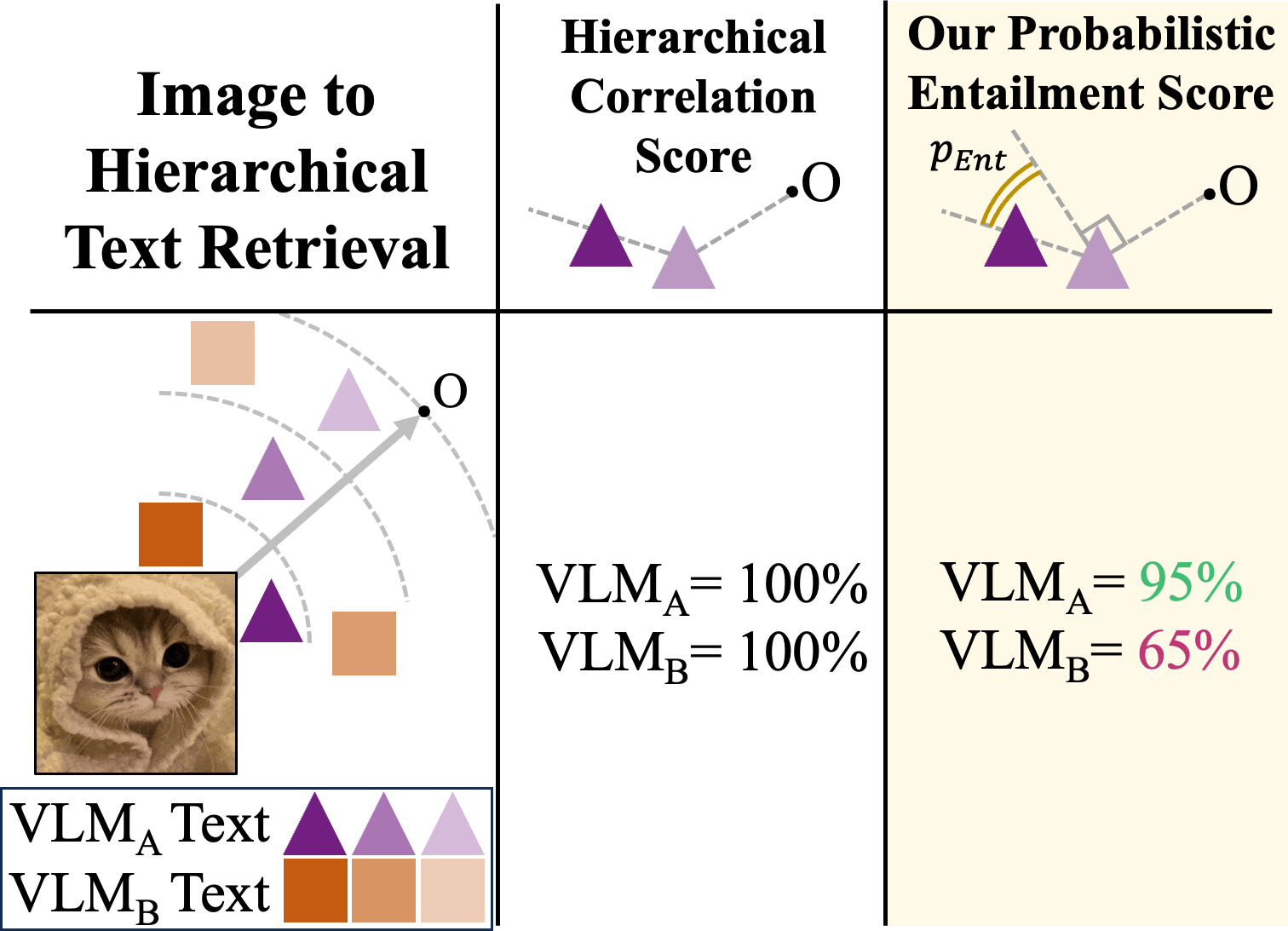}
        \vspace{-0.2cm}
        \vfill
        {\small (b) Fine-grained PEP evaluation.}
    \end{minipage}

    \vspace{0.3cm} 
    \caption{
        \textbf{\methodName improves both hierarchical training and evaluation.} 
        \textbf{(a)} \methodName outperforms the HyCoCLIP baseline on downstream tasks and our hierarchical benchmark (PEP AUC). 
        \textbf{(b)} Our new \textbf{Probabilistic Entailment Score} ($p_{\text{Ent}}$) offers a more discriminative evaluation; while both models may achieve 100\% correlation, our metric correctly identifies $VLM_A$ as superior.
    }
    \label{fig:teaser}
    \vspace{-0.5cm}
\end{figure}

\begin{abstract}

Large-scale Vision–Language Models (\vlms) such as \clip learn powerful semantic representations but operate in Euclidean space, which fails to capture the inherent hierarchical structure of visual and linguistic concepts. Hyperbolic geometry, with its exponential volume growth, offers a principled alternative for embedding such hierarchies with low distortion. 
However, existing hyperbolic VLMs use entailment losses that are unstable: as parent embeddings contract toward the origin, their entailment cones widen toward a half-space, causing catastrophic cone collapse that destroys the intended hierarchy. 
Additionally, hierarchical evaluation of these models remains unreliable, being largely retrieval-based and correlation-based metrics and prone to taxonomy dependence and ambiguous negatives.
To address these limitations, we propose an adaptive entailment loss paired with a norm regularizer that prevents cone collapse without heuristic aperture clipping. 
We further introduce an angle-based probabilistic entailment protocol (PEP) for evaluating hierarchical understanding, scored with AUC-ROC and Average Precision. 
This paper introduces a stronger hyperbolic VLM baseline \textbf{ARGENT}, \textbf{A}daptive hie\textbf{R}archical ima\textbf{G}e-t\textbf{E}xt represe\textbf{NT}ation.
\methodName improves the \sota hyperbolic \vlm by $0.7$, $1.1$, and $0.8$ absolute points on image classification, text-to-image retrieval, and proposed hierarchical metrics, respectively.


\end{abstract}

\renewcommand{\thefootnote}{$\dag$}
\footnotetext[0]{Work done while at Samsung Research America.}
\renewcommand{\thefootnote}{\arabic{footnote}}
\section{Introduction}
\label{sec:intro}

Large-scale Vision-Language Models (VLMs) like \clip~\cite{radford2021learning, jia2021scaling} have revolutionized the field by learning semantically aligned representations from web-scale image-text pairs. Their success, however, rests on a ``flat-world'' assumption: concepts are represented as points in Euclidean space, implicitly treating all semantic relationships as uniform distances. This clashes with the deeply hierarchical structure of both the visual world and human language \cite{desai2023hyperbolic}, where a \textit{``poodle''} is a specific type of \textit{``dog''}, which is a class of \textit{``animal''}. While recent work reports traces of emergent hierarchical structure in \clip-like models \cite{alper2024emergent,chunprobabilistic,chou2024embedding}, these effects are inconsistent, dataset-dependent, and difficult to control.
The primary bottleneck often appears to be the text encoder, which flattens structured language into a single vector. This approach struggles to explicitly model linguistic hierarchy, hindering its alignment with structured visual information. For example, the text encoder may represent \textit{``a poodle''} and \textit{``a dog''} as closely related, yet fail to encode the explicit `is-a' relationship, treating it similarly to a purely associative link (like \textit{``dog''} and \textit{``leash''}).

The core challenge in visual-language modeling (VLM) is geometrically accommodating the inherent exponential complexity of the conceptual and compositional hierarchies (\eg, \textit{``dog''} $\rightarrow$ \textit{``mammal''} $\rightarrow$ \textit{``animal''}). Embedding this structure into a traditional Euclidean space, which exhibits only polynomial volume growth, is not only mathematically inefficient but fundamentally induces high distortion. To resolve this, recent work~\cite{desai2023hyperbolic,ramasinghe2024accept,PalSDFGM2024} has leveraged hyperbolic geometry. This non-Euclidean space is defined by its constant negative curvature and, crucially, its exponential volume growth with respect to the radius. This property makes it the continuous analog of a discrete tree and a natural, low-distortion substrate for hierarchical data.

The application of hyperbolic geometry to VLMs has progressed rapidly. Initial explorations, such as MERU~\cite{desai2023hyperbolic}, pioneered contrastive learning in hyperbolic space. Subsequent efforts, like HyCoCLIP~\cite{PalSDFGM2024}, advanced this by modeling compositional relationships using hyperbolic entailment cones. While foundational, existing hyperbolic VLMs face two significant challenges that impede further progress. On the methodological front, their core loss function suffers from \textbf{Entailment Loss Instability}, making training brittle and prone to collapse. Separately, the evaluation framework for these models exhibits a critical shortcoming: current hierarchical reasoning benchmarks are often unreliable and sensitive to false negatives, failing to accurately measure true hierarchical understanding.

We first address the critical challenge of \textbf{Entailment Loss Instability}. The entailment loss in prior methods is inherently unstable due to a geometric vulnerability. The hierarchical relationship is defined by a cone whose aperture is inversely coupled to the norm of the parent embedding. The model can exploit this coupling to find a pathological solution: by minimizing the parent norms, the cone's aperture widens until it degenerates into a flat half-space. This collapse destroys the geometric notion of entailment, leading to a collapsed representation space and catastrophic training instability. HyCoCLIP~\cite{PalSDFGM2024} attempted to mitigate this with a heuristic regularizer, $\eta$, to manually restrict the cone's aperture. However, this is an ad-hoc fix that fails to address the fundamental geometric cause.

We address this by proposing the Adaptive Entailment (AdaEnt) Loss that dynamically scales the entailment angle based on embedding similarity, removing the detrimental norm-to-cone relationship without relying on unstable heuristics. We stabilize it further by integrating AdaEnt with a norm regularizer. \Fref{fig:teaser}\textcolor{cvprblue}{a} shows that our resulting model, ARGENT, trained with the adaptive loss, outperforms the HyCoCLIP baseline by a large margin.


Second, we tackle the pervasive problem of \textbf{Inadequate Hierarchical Evaluation}. Prior benchmarks, often relying on WordNet, are limited to short, single-level phrases. While more recent datasets like HierarCaps~\cite{alper2024emergent} provide valuable multi-level captions, their established evaluation protocol suffers from two issues: (1) The inherent hierarchical signal within these datasets is weak due to ambiguous ordering. (2) More critically, the evaluation procedure introduces a systemic bias by incorrectly defining the ground truth. During standard image-to-text retrieval, only the single caption paired with the query image is considered a positive match. This ignores that many other captions in the candidate pool could be equally valid hierarchical parents or children, preventing the evaluation from measuring true hierarchical generalization.


To address this, we propose a new, direct, and robust Probabilistic Entailment Protocol (PEP). Leveraging HierarCaps, we redefine the metric by treating the angle between embeddings as a direct proxy for entailment probability. We then compute AUC-ROC and Average Precision (AP) across a carefully selected set of intra- and inter-sample pairs. This protocol moves beyond brittle retrieval accuracy to provide a significantly more reliable and fine-grained measure of a model’s hierarchical understanding. 
\Fref{fig:teaser}\textcolor{red}{b} shows that our metric provides a fine-grained score that correctly distinguishes model performance where the original protocol fails.

In summary, this paper makes three key contributions to advance how VLMs represent conceptual hierarchies:
\begin{itemize}
    \item We introduce the novel Adaptive Entailment (AdaEnt) Loss, which resolves the critical Entailment Loss Instability in hyperbolic VLMs, ensuring stable training and robust representation (\cref{sec:method}).
    \item We propose the Probabilistic Entailment Protocol (PEP), a new, objective evaluation that overcomes the systematic biases of prior benchmarks to reliably measure hierarchical generalization (\cref{sec:revisit_evaluation}).
    \item Through comprehensive analysis, we validate our proposed loss and protocol, demonstrating that \methodName sets a new state-of-the-art on hierarchical representation learning benchmarks (\cref{sec:experiments}).
\end{itemize}

Together, these contributions provide a stronger foundation for building and validating the next generation of hyperbolic VLMs.
\section{Related Work}

This work builds upon advancements in three primary areas: large-scale vision-language models, the application of hyperbolic geometry to deep learning, and the rapidly emerging field of hyperbolic vision-language models.

\mypara{Vision-Language Models}
Large-scale Vision-Language Models (\vlms), pre-trained on extensive web data, have emerged as the leading approach in multi-modal research. 
Models such as \clip~\cite{radford2021learning} and ALIGN~\cite{jia2021scaling} utilize a contrastive loss on massive image-text datasets to learn a joint, semantically rich embedding space. 
This shared space shows remarkable zero-shot transfer abilities across various tasks. 
However, the inherent Euclidean geometry of these models is not optimal for representing data with implicit hierarchical structures, such as semantic taxonomies. 
This paper extends these foundational architectures by adopting an alternative geometry specifically tailored to model such structures more effectively.
\vspace{-0.1cm}

\mypara{Hyperbolic Deep Learning}
Hyperbolic geometry, a non-Euclidean space with constant negative curvature, is promising for embedding hierarchical data~\cite{nickel2017poincare, chamberlain2017neural}. 
Early work in natural language processing~\cite{dhingra2018embedding, tifrea2018poincar} and graph representation learning~\cite{liu2019hyperbolic, franco2023hyperbolic,flaborea2024contracting} demonstrated the advantage of hyperbolic embeddings for representing taxonomies like WordNet~\cite{miller1995wordnet} and large-scale graph structures. 
These methods commonly utilize the Poincaré ball model to adapt neural network architectures to this curved space. 
This work applies their geometric principles to learn representations for multi-modal data.
\vspace{-0.1cm}

\mypara{Hyperbolic Vision-Language Models}
The success of hyperbolic geometry in unimodal tasks has led to its recent adoption in vision-language modeling. Early efforts~\cite{kim2024hype,yang2024hypformer}, such as \meru~\cite{desai2023hyperbolic}, extended \clip{}'s contrastive learning framework to the Poincaré ball, introducing hyperbolic \vlms. 
\hycoclip~\cite{PalSDFGM2024} advanced this by explicitly enforcing compositional and hierarchical relationships using hyperbolic entailment cones, moving beyond basic contrastive learning. 
ATMG~\cite{ramasinghe2024accept} explored the modality gap in hyperbolic space, proposing mitigation strategies for this issue, often worsened by the space's curvature. Recent works~\cite{yang2024hyperbolic,mandica2024hyperbolic,chen2024hyperbolic} applies the hyperbolic concept to vision large language models or extends to work video modality~\cite{li2025enhancing,li2025hlformer,wen2025hover} or adapts to specific tasks~\cite{shimizu2024fashion,qiu2024hihpq,gonzalez2025hyperbolic}.
This work complements these efforts by focusing on training stability and evaluation of hierarchical structures within the embedding space, rather than solely on the initial separation of modalities.
\vspace{-0.1cm}

\mypara{Evaluation of Hierarchical Reasoning}
Existing methods for evaluating hierarchical reasoning often rely on suboptimal benchmarks. WordNet-based evaluations ~\cite{PalSDFGM2024} are typically restricted to simple, single-level relationships and depend on a single, manually curated taxonomy. 
While more recent datasets like HierarCaps~\cite{alper2024emergent} offer multi-level captions, their evaluation protocol has fundamental flaws. 
Specifically, conventional retrieval-based metrics fail to account for valid hierarchical relationships with items outside of the query's ground-truth pairs, thus hindering the accurate measurement of a model's true generalization capabilities. 
This work directly addresses these limitations by introducing a novel and more robust evaluation protocol designed to precisely assess a model's comprehension of hierarchical entailment.
\vspace{-0.1cm}

\section{Background}
This section provides the necessary background on the Lorentz model of hyperbolic geometry and its application to hierarchical representation learning.

\begin{figure}[t]
    \centering
    \vspace{-0.3cm}
    \includegraphics[width=0.6\linewidth]{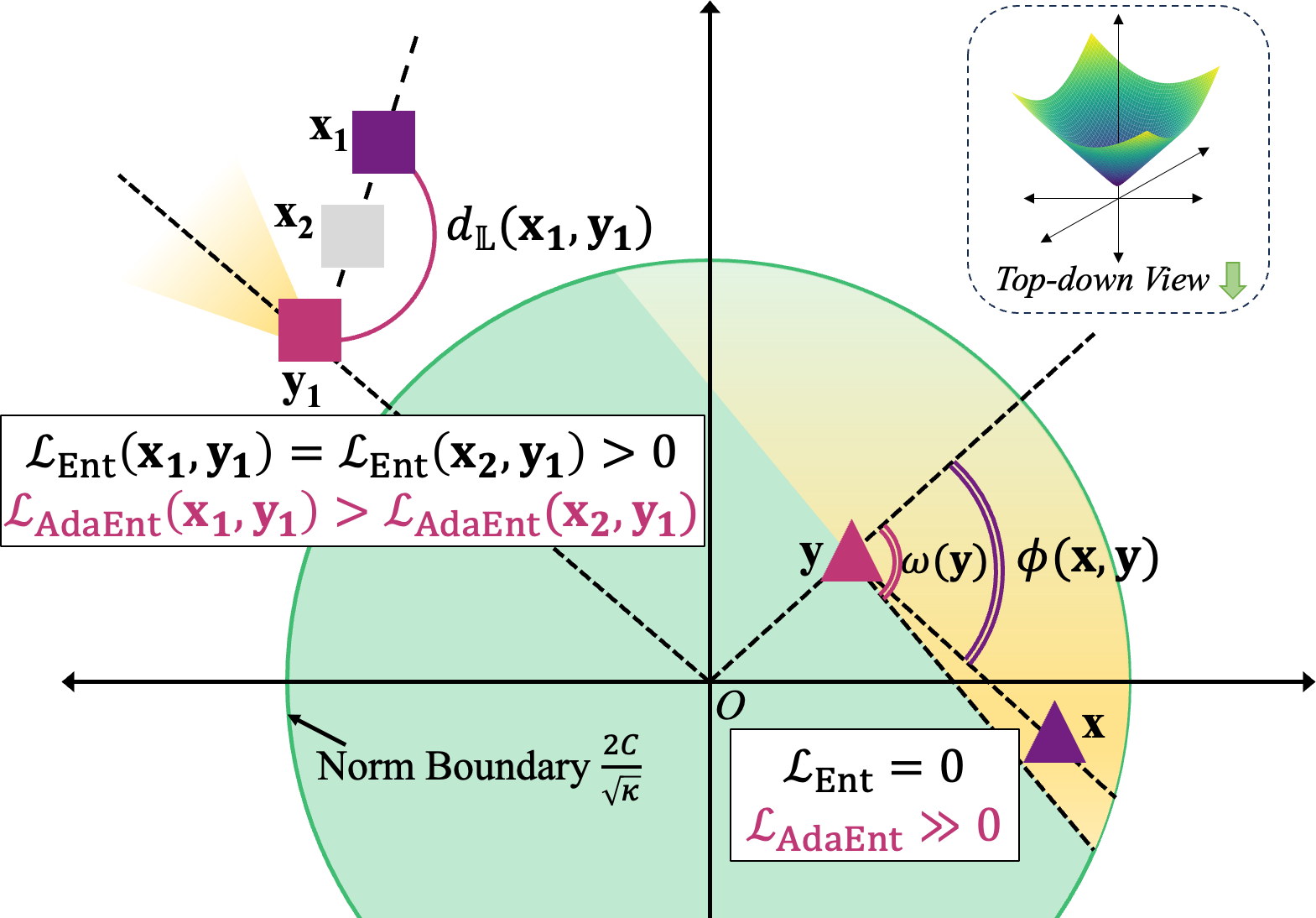}
    \caption{\textbf{Behavior of Adaptive Entailment $\mathcal{L}_\text{AdaEnt}$ and standard Entailment Loss $\mathcal{L}_\text{Ent}$.} The figure highlights two cases in a top-down view of hyperbolic space: \textbf{(1)} \textbf{Inside the norm boundary} ($\|\tilde{\mathbf{y}}\| \le \frac{2C}{\sqrt{\kappa}}$): The standard $\mathcal{L}_\text{Ent}$ collapses to zero for all $\mathbf{x}$ in the non-origin half-space, even when the exterior angle $\phi(\mathbf{x},\mathbf{y})$ is large. Our $\mathcal{L}_\text{AdaEnt}$ remains active ($\mathcal{L}_\text{AdaEnt} \gg 0$), preventing vanishing gradients. \textbf{(2)} \textbf{Outside the norm boundary}: $\mathcal{L}_\text{Ent}$ penalizes the likely noisy positive $\mathbf{x}_2$ and the true negative $\mathbf{x}_1$ with the same value. Our $\mathcal{L}_\text{AdaEnt}$ adaptively assigns a lower loss to $\mathbf{x}_2$ while strongly penalizing $\mathbf{x}_1$.}
    \label{fig:notation}
    \vspace{-0.5cm}
\end{figure}

\begin{figure*}[t]
    \vspace{-0.3cm}
    \centering
    \begin{subfigure}[t]{0.4\linewidth}
        \includegraphics[width=\linewidth]{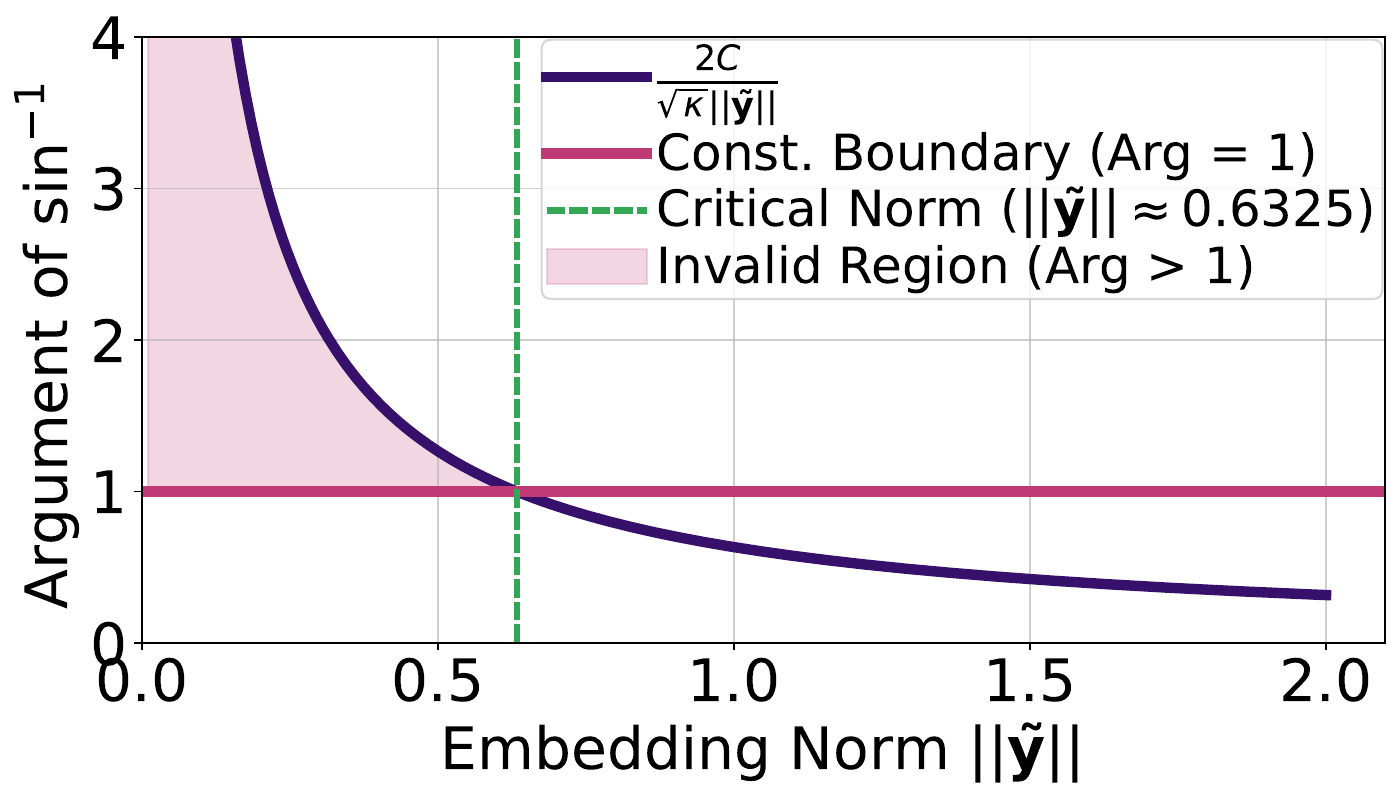}
        \caption{}
        \label{fig:method_ent_constraint}
    \end{subfigure}
    \hspace{0.5cm}
    \begin{subfigure}[t]{0.4\linewidth}
        \includegraphics[width=\linewidth]{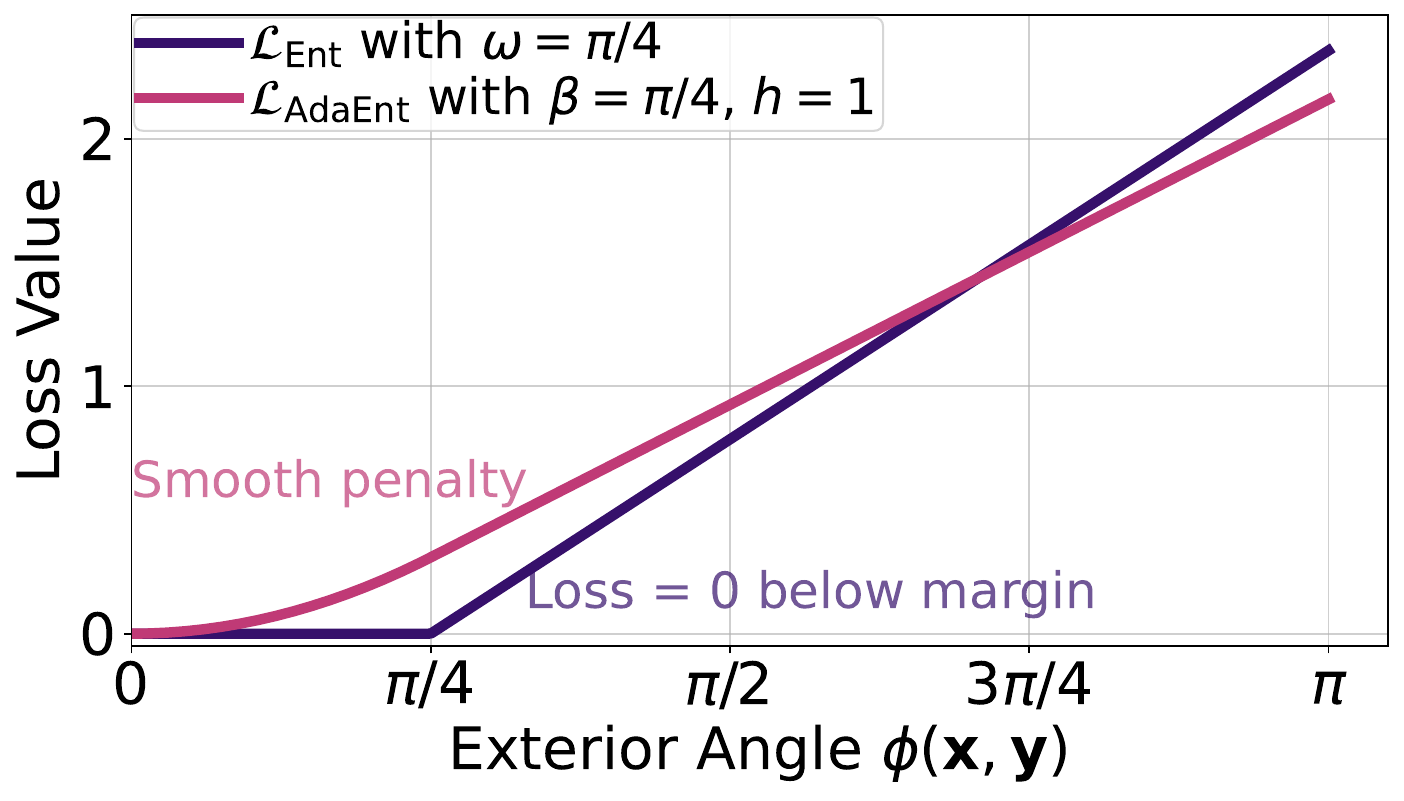}
        \caption{}
        \label{fig:method_ent_comparison}
    \end{subfigure}
    \begin{subfigure}[t]{0.4\linewidth}
        \includegraphics[width=\linewidth]{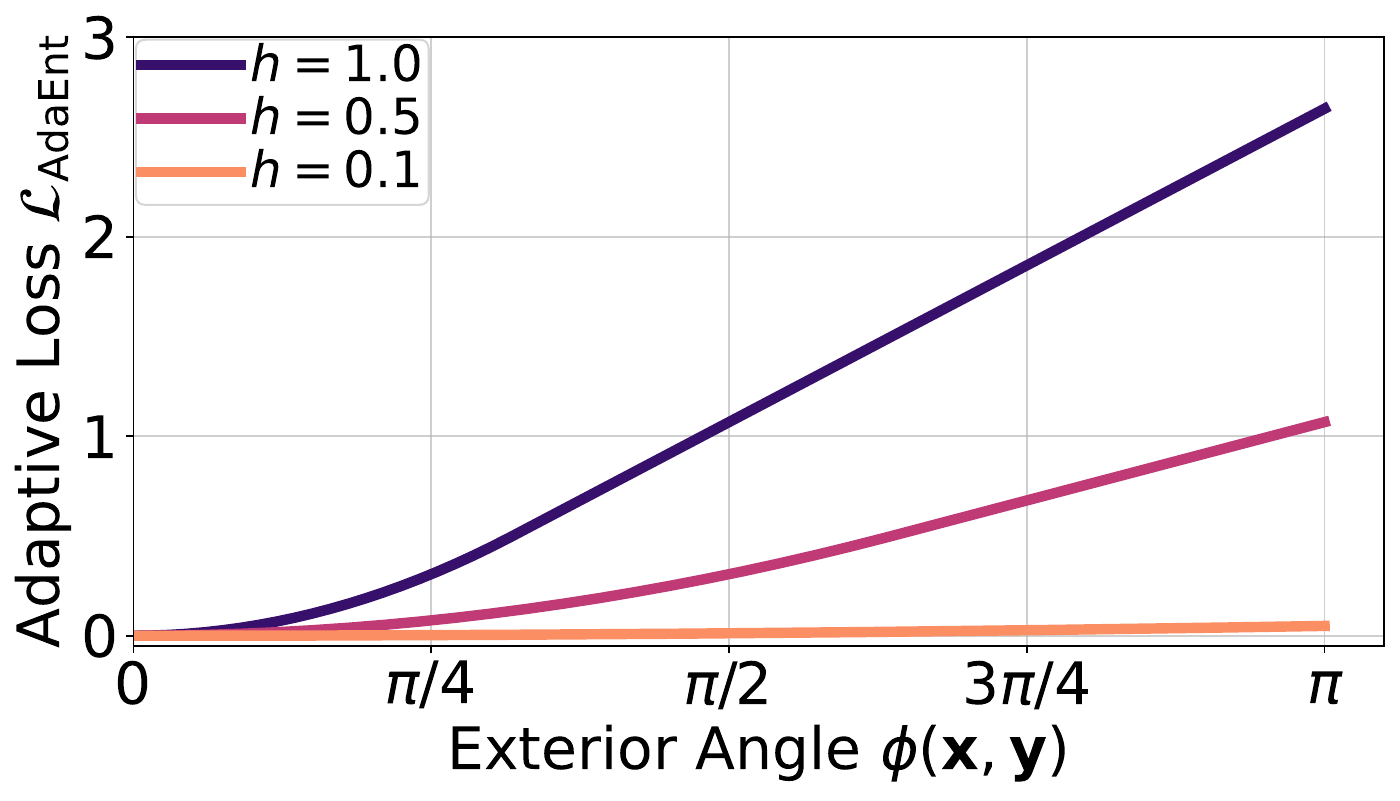}
        \caption{}
        \label{fig:method_ent_adaptive}
    \end{subfigure}
    \hspace{0.5cm}
    \begin{subfigure}[t]{0.4\linewidth}
        \includegraphics[width=\linewidth]{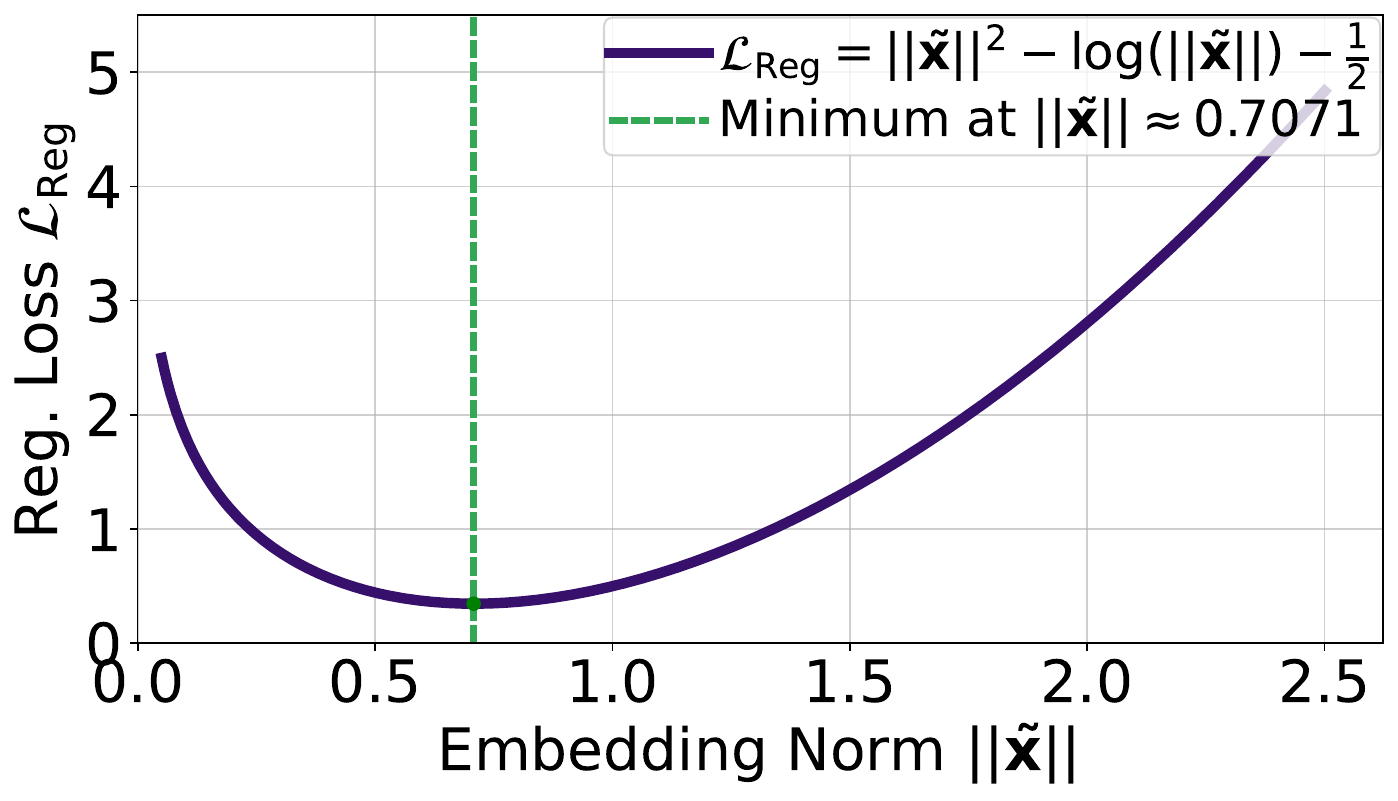}
        \caption{}
        \label{fig:method_ent_reg}
    \end{subfigure}
    \vspace{-0.2cm}
    \caption{
    \textbf{Analysis of the vanilla and 
    proposed adaptive entailment loss}. (a) The norm constraint required by the standard loss. (b) A comparison between loss functions. (c) The effect of our adaptive weight. (d) The behavior of our norm regularizer.}
    \vspace{-0.5cm}
\end{figure*}

\mypara{Hyperbolic Representations in the Lorentz Model} 
To effectively embed data with a tree-like structure, \meru \cite{desai2023hyperbolic} utilize hyperbolic geometry, which offers a powerful inductive bias due to its constant negative curvature and exponential volume growth. 
\meru selects the Lorentz model (or the hyperboloid model), a well-established framework for hyperbolic geometry. 
This model defines the $n$-dimensional hyperbolic space $\Lo^n$ as a specific manifold within the $(n+1)$-dimensional Minkowski spacetime, $\R^{n+1}$. 
The Lorentzian inner product governs the geometry of this spacetime, defined for vectors $\mathbf{x},\mathbf{y}\in\R^{n+1}$ as:
\begin{align}
    \langle \mathbf{x}, \mathbf{y}\rangle_\Lo = -x_0y_0 + \langle \mathbf{\tilde{x}}, \mathbf{\tilde{y}} \rangle_\mathbb{E}
\end{align}
where $x_0, y_0$ are the first \textit{time} coordinates and $\mathbf{\tilde{x}}, \mathbf{\tilde{y}}$ are the remaining $n$ \textit{spatial} coordinates. The hyperbolic manifold $\Lo^n$ is then the set of points on the upper sheet of a two-sheeted hyperboloid:
\begin{align}
    \Lo^n = \left\{ \mathbf{x} \in \R^{n+1}: \langle \mathbf{x}, \mathbf{x} \rangle_\Lo= -\frac{1}{\kappa}, \kappa > 0 \right\}
\end{align}
Here, $-\kappa\in\R$ represents the curvature of the space. For any point $\mathbf{x}$ on this manifold, its time coordinate is fully determined by its spatial coordinates via the relation $x_0 = \sqrt{1/\kappa + \|\mathbf{\tilde{x}}\|^2}$. The distance between two points is measured along the geodesic connecting them, calculated as:
\vspace{-0.4cm}
\begin{align}
    d_\Lo (\mathbf{x}, \mathbf{y}) = \frac{1}{\sqrt{\kappa}} \cosh^{-1} \left( -\kappa \langle \mathbf{x}, \mathbf{y} \rangle_\Lo \right).
\end{align}
Vanilla deep learning encoders produce vectors in Euclidean space. 
To map these vectors onto the hyperbolic manifold, \meru predicts in the tangent space at the origin of $\Lo^n$ and projects them using the exponential map, $\exp^\kappa_\mathbf{o}$.

\mypara{Learning Alignment and Hierarchy}
Prior works \cite{desai2023hyperbolic,PalSDFGM2024} focus on two primary objectives: aligning related concepts (\eg an image of a cat and the text ``a photo of a cat'') and hierarchical ordering (\eg ``cat'' is a type of ``animal'', the crop of cat region is more generic than the full image of a cat and a dog).
These methods align image-text via contrastive learning. Given a batch $B$ of image-text pairs $(I, T)$, a standard contrastive loss function $\mathcal{L}_\text{Cont}$ pulls corresponding representations together while pushing non-corresponding pairs apart:
\begin{align}
    \mathcal{L}_\text{Cont} (I, T) = - \sum_{i\in B} \log \frac{\exp\left( s(I_i, T_i) / \tau\right)}{\sum\limits_{j\in B} \exp(s(I_i, T_j)/\tau)},
\end{align}
where the similarity score $s(\mathbf{x}, \mathbf{y})$ is the negative geodesic distance, $s(\mathbf{x}, \mathbf{y}) = -d_\Lo(\mathbf{x}, \mathbf{y})$ and $\tau$ is a learnable temperature parameter.

To explicitly model ``is-a'' relationships, \hycoclip \cite{PalSDFGM2024} utilizes a geometric construct of an entailment cone.
For any given concept embedding $\mathbf{y}$, its entailment cone $\Lambda_q$ defines a region in the space. The learning objective is to position embeddings of more specific concepts (\eg \textit{``poodle''}, $\mathbf{x}$) inside the cones of more general concepts (\eg \textit{``dog''}, $\mathbf{y}$). The size of this cone is determined by its half-aperture, $\omega(\mathbf{y})$:
\vspace{-0.4cm}
\begin{align}
    \omega(\mathbf{y}) = \sin^{-1} \left( \frac{2C}{\sqrt{\kappa}\|\mathbf{\tilde{y}}\|}\right),
    \label{eq:aper}
\end{align}
where the constant $C{=}0.1$ \cite{PalSDFGM2024} keeps the embedding closer to the origin.
This formulation is position-dependent; semantically more general concepts that lie closer to the origin (smaller $\|\mathbf{\tilde{y}}\|$) are assigned wider cones. 
An entailment loss, $\mathcal{L}_\text{ent}$, enforces this structure by penalizing cases where a specific concept $\mathbf{x}$ lies outside the cone of a general concept $\mathbf{y}$:
\vspace{-0.4cm}
\begin{align}
    \mathcal{L}_\text{Ent}(\mathbf{x}, \mathbf{y}) = \max (0, \phi(\mathbf{x}, \mathbf{y}) - \eta\omega(\mathbf{y})) \label{eq:ent}
\end{align}
where $\phi(\mathbf{x}, \mathbf{y}) = \pi - \angle \mathbf{Oyx}$ is the exterior angle of $\mathbf{x}$ and $\eta$ is a scaling factor~\cite{PalSDFGM2024}. Prior works~\cite{PalSDFGM2024} apply this loss to various pairs, such as \textit{(image, text)}, \textit{(image crop, full image)}, and \textit{(text phrase, full caption)}.
We show many of these concepts pictorially in \cref{fig:notation}.

\section{\methodName}
\label{sec:method}

We propose a novel adaptive entailment learning method to improve the alignment of latent representations in hyperbolic space. Specifically, we introduce our adaptive entailment loss, which is designed to overcome the issues of cone collapse and instability in prior work, thereby facilitating significantly more robust hierarchical learning.




A core issue with the standard entailment loss $\mathcal{L}_\text{Ent}$ in \Eref{eq:ent}, which hinders learning meaningful representations, stems from the definition of the half-aperture in \Eref{eq:aper}. 
The $\sin^{-1}$ function is only defined if its argument is $\le 1$, which imposes the constraint $\|\mathbf{\tilde{y}}\| \ge \frac{2C}{\sqrt{\kappa}}$. 
With typical converged values of $C{=}0.1$ and $\kappa{=}0.1$ \cite{PalSDFGM2024}, this requires $\|\mathbf{\tilde{y}}\| \ge 0.6325$. \Fref{fig:method_ent_constraint} illustrates this invalid domain (shaded region). 
However, many embeddings in previous works~\cite{PalSDFGM2024,desai2023hyperbolic} fall into the invalid region, violating this constraint. In such cases, the aperture is clipped to $\pi/2$, transforming the cone into a half-space and reducing the entailment objective to a simple norm constraint: $\|\mathbf{\tilde{y}}\|{\le} \|\mathbf{\tilde{x}}\|$. 
The hyperparameter $\eta$ in \Eref{eq:ent} of \hycoclip \cite{PalSDFGM2024} thus serves as an ad-hoc fix, manually tuning cone sizes to manage these geometric inconsistencies.


\begin{figure}[!tb]
    \vspace{-0.3cm}
    \centering
    \includegraphics[width=0.45\linewidth]{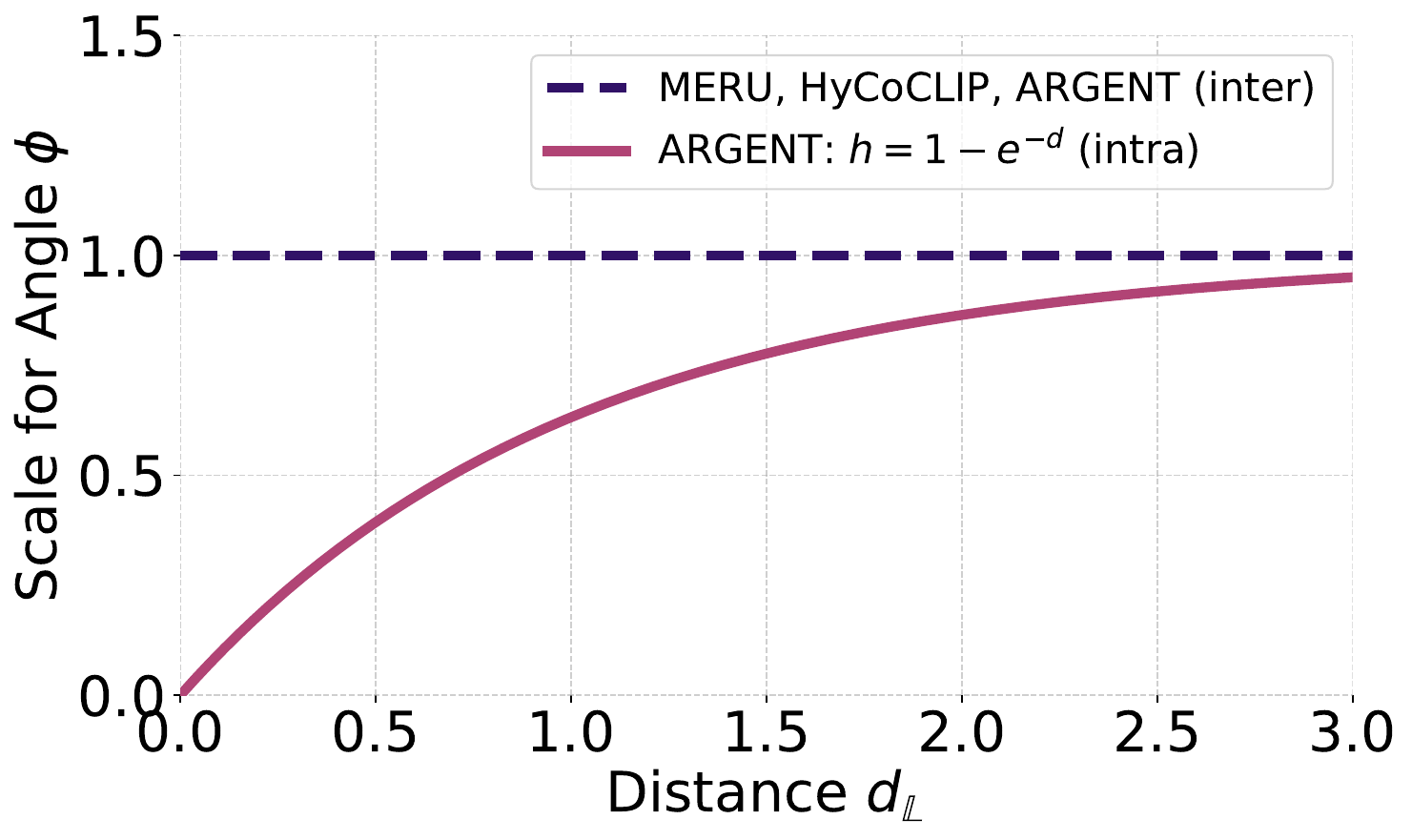}
    \vspace{-0.2cm}
    \caption{\textbf{Weighing factor $h(\mathbf{x}, \mathbf{y})$ (scaling) \wrt the distance} (negative of similarity) for intra and inter-modality samples in \meru, \hycoclip and \methodName. 
    \textbf{\methodName uses adaptive weights} compared to constant weights in \meru and \hycoclip.}
    \label{fig:method_angle_scale}
    \vspace{-0.5cm}
\end{figure}

To address this issue, we introduce a more direct and robust \textbf{\textit{Adaptive Entailment Loss}}. For a specific concept $\mathbf{x}$ and a general concept $\mathbf{y}$, we formulate the loss $\mathcal{L}_\text{AdaEnt}$ as:
\begin{align}
    \mathcal{L}_\text{AdaEnt} = \text{Huber} \left( h(\mathbf{x}, \mathbf{y}) \cdot \phi (\mathbf{x}, \mathbf{y}), \beta\right)
\end{align}
where $\beta$ is the Huber loss delta. \Fref{fig:method_ent_comparison} shows that this formulation directly minimizes the exterior angle and provides a smoother loss surface than $\mathcal{L}_\text{ent}$, without relying on the problematic aperture calculation.

The adaptive weight term $h(\mathbf{x}, \mathbf{y})$ modulates the loss strength, with its effect visualized in \Fref{fig:method_ent_adaptive}. For inter-modality pairs (\eg image-text), we set $h(\mathbf{x}, \mathbf{y}) = 1$. For intra-modality pairs (\eg a crop and its full image), we define:
\begin{align}
    h(\mathbf{x},\mathbf{y}) &= 1{-}\exp (s(\mathbf{x},\mathbf{y})) = 1{-}\exp (-d_\Lo(\mathbf{x},\mathbf{y}))
\end{align}
This adaptive weighting prevents the entailment loss from pushing near-identical pairs apart, such as when a crop is almost identical to the full image.
Thus, the adaptive loss removes explicit geometric constraints of the entailment cone.
\cref{fig:method_angle_scale} visualizes the adaptive term $h(\mathbf{x}, \mathbf{y})$ against the distance $d_\Lo(\mathbf{x},\mathbf{y})$ across models.

While improving stability, it drifts the embedding away from the origin, creating an overly sparse representation space. To counteract this, we introduce a regularization term:
    $\mathcal{L}_\text{Reg} (\mathbf{x}) = \|\mathbf{\tilde{x}}\|^2{-}\log ( \|\mathbf{\tilde{x}}\|).$
\Fref{fig:method_ent_reg} shows that this regularizer encourages embedding norms to remain within a stable range by penalizing values that are either too small or too large.
Our training objective is a combination of the contrastive, adaptive entailment, and regularization losses:
$\mathcal{L} = \mathcal{L}_\text{Cont} + \gamma_1 \mathcal{L}_\text{AdaEnt} + \gamma_2 \mathcal{L}_\text{Reg}$
where, $\gamma_1,\gamma_2$ are the weighing terms.
\vspace{-0.3cm}
\section{Revisiting Hierarchical Entailment Evaluation}\label{sec:revisit_evaluation}
To our knowledge, HierarCaps~\cite{alper2024emergent} is the only benchmark for evaluating multi-level, hierarchical image-text representations. It provides semi-automatically generated captions at varying detail levels to evaluate entailment. However, we identify two critical, previously undiscussed limitations in its evaluation protocols:

\mypara{Insensitivity of Ranking-Based Metrics} 
The primary ranking metric, Kendall's Correlation, is insensitive to the magnitude of ordering errors. For example, consider four captions ordered from coarse to fine: \textit{``people''; ``a group of people'', ``a group of people milling about inside a large space.''; ``a group of people milling about inside a large space in the winter''.} This metric assigns the same penalty for a swap between the two coarsest levels (\eg \textit{``people''} and \textit{``a group of people''}) as it does for a swap between the two finest levels. Intuitively, the latter error is more severe as the captions differ by a greater amount of semantic detail. Furthermore, the metric fails to capture the degree of entailment. If two models produce the same correct ordering, Kendall's correlation cannot differentiate which model better represents the semantic hierarchy.

\mypara{Ambiguity in Retrieval-Based Evaluation} Image-to-text retrieval protocols use a global candidate pool that mixes true negatives with many false negatives. We observe cases where captions from visually similar images provide (near) valid descriptions, yet evaluation counts them as distractors and penalizes them. We perform two analyses to quantify this impact.

\begin{figure}[!t]
    \centering
    \includegraphics[width=0.45\linewidth]{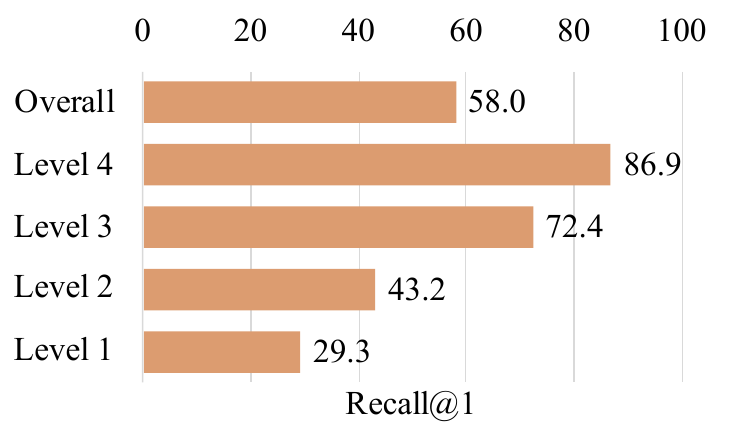}
    \vspace{-10pt}
    \caption{\textbf{Recall@1} of the BLIP-L model in image-to-text retrieval task on HierarCaps. Level 1 and 4 denotes the most generic and the most specific captions. We restrict the candidate pool to a local set containing ground-truth captions and the top-5 most similar (presumed false negative) captions. \textbf{The model performance degrades when the text is more generic (Level 1).}}
    \label{fig:benchmark_analysis1}
\end{figure}

\begin{table}[t]
    \centering
    \caption{\textbf{HierarCaps image-to-text retrieval performance of $\text{\clip}^\text{L}_\text{FT}$} increases as false negative captions are progressively removed from the candidate pool. This shows that the current evaluation protocol is flawed.}
    \label{tab:benchmark_analysis2}
    \vspace{-0.2cm}
    \footnotesize
    \begin{tblr}{width=0.6\linewidth,colspec={@{}X[1,l]X[1,l]X[l]@{}},stretch=0}
    \toprule
     Remove top-k & Precision (\%) $\uparrowRHDSmall$ & Recall (\%) $\uparrowRHDSmall$ \\
     \hline
     None & 15.36 & 43.23 \\
     k=5 & 17.10~\pdif{1.74} & 46.45~\pdif{3.22} \\
     k=10 & 18.05~\pdif{2.69} & 48.10~\pdif{4.87} \\
     \bottomrule
    \end{tblr}
\end{table}

In our first analysis, we use a BLIP-L-ITM~\cite{li2022blip} model finetuned on \coco~\cite{lin2014microsoft} to retrieve the top-5 most similar images for each image in the benchmark. For every image, we construct a local candidate pool consisting of its ground-truth captions and the captions from these 5 similar images. We then re-evaluate the BLIP model on image-to-text retrieval at each hierarchical level using only this local pool. \Fref{fig:benchmark_analysis1} shows the Recall@1 from the most generic (Level 1) to the most specific (Level 4). At the most generic level, performance drops to 30\%. Although we expect some overlap for generic concepts, this sharp degradation within a small, highly confounding local pool indicates that the original retrieval protocol defines an unreliable evaluation task.

In our second analysis, we evaluate $\text{\clip}^\text{L}_\text{FT}$~\cite{alper2024emergent}, a model fine-tuned on hierarchical data, using cleaner global candidate pools. This avoids potential artifacts from the BLIP model, which was not trained on hierarchical data. We follow the original HierarCaps protocol but progressively remove captions from the top-$k$ similar images (identified by BLIP) from the global pool. \Tref{tab:benchmark_analysis2} shows that retrieval performance steadily increases as we remove more of these ambiguous false negatives. This confirms the highly significant and negative impact of dataset ambiguity on the retrieval evaluation.

To reduce the impact of noisy and ambiguous annotations, we construct a cleaned benchmark by filtering out ambiguous image–caption pairs and retaining only high-confidence matches. We apply a model-assisted cleaning pipeline based on Qwen3-VL~\cite{bai2025qwen3} (verification), Qwen3-VL-Embed and Qwen3-VL-Reranker~\cite{li2026qwen3} (similarity); details and thresholds are provided in the supplementary. From 1,000 samples, we retain 522 image–caption pairs with strong alignment and low ambiguity. We also build a negative caption pool by selecting 100 captions with the lowest retrieval degree (i.e., rarely retrieved for other images).

\mypara{Proposed Probabilistic Entailment Protocol} 
We propose a new protocol to evaluate hierarchical entailment on the HierarCaps dataset. Our metric quantifies the entailment probability $p_\text{Ent}$ via the angle $\phi$ between embeddings:
\begin{align}
    p_\text{Ent}(\mathbf{x}, \mathbf{y}) = \max \left(1 - \frac{2\phi(\mathbf{x}, \mathbf{y})}{\pi}, 0\right)\label{eq:p_ent}
\end{align}
This function maps a perfect alignment ($\phi=0$) to $p_\text{Ent}=1$ and an orthogonal or opposed alignment ($\phi \ge \pi / 2$) to $p_\text{ent}=0$. For each image, we compute $p_\text{Ent}$ using all corresponding hierarchical captions as positives, and use fine-grained captions from the non-ambiguous negative pool as negatives. We then report the Area Under the Receiver Operating Characteristic Curve (AUC-ROC) and Average Precision (AP) as the final metrics for quantifying entailment quality.

\begin{table*}[!t]
    \centering
    \caption{\textbf{Main Results.} 
    \textbf{\methodName outperforms baselines on almost all metrics}, often by a large margin. The adaptive approach not only improves performance on downstream tasks (classification and retrieval) but also reduces hierarchical errors. 
    Results of all 16 image classification datasets are in supplementary material.
    [Keys: \textbf{Best}, \inc{Gain} and \dec{drop} relative to \hycoclip baseline \cite{PalSDFGM2024}, Cls= Classifcation, T2I= Text-to-Image, I2T= Image-to-Text, Ret= Retrieval]}
    \label{tab:quan}
    \scalebox{0.7}{
    \setlength{\tabcolsep}{0.05cm} 
    \begin{NiceTabular}{l @{\hspace{12pt}} cc @{\hspace{12pt}} cc @{\hspace{6pt}} cc @{\hspace{12pt}} cc @{\hspace{6pt}} cc @{\hspace{12pt}} ccc @{\hspace{6pt}} cc}
        \toprule
        \multirow{3}{*}{\textbf{Model}} & \multicolumn{2}{c}{\textbf{Cls (Acc $\uparrowRHDSmall$)}} & \multicolumn{4}{c}{\textbf{T2I Ret (R@k $\uparrowRHDSmall$)}} & \multicolumn{4}{c}{\textbf{I2T Ret (R@k $\uparrowRHDSmall$)}} & \multicolumn{5}{c}{\textbf{Hierarchical Metrics}} \\
        \cmidrule[0.7pt](r{2pt}){2-3} \cmidrule[0.7pt](lr{16pt}){4-7} \cmidrule[0.7pt](lr{16pt}){8-11} \cmidrule[0.7pt](lr){12-16}
        & \multirow{2}{*}{INet} & \multirow{2}{*}{Avg.} & \multicolumn{2}{c}{COCO} & \multicolumn{2}{c}{Flickr} & \multicolumn{2}{c}{COCO} & \multicolumn{2}{c}{Flickr} & \multicolumn{3}{c}{WordNet} & \multicolumn{2}{c}{PEP} \\
        \cmidrule(lr){4-5} \cmidrule(lr){6-7} \cmidrule(lr){8-9} \cmidrule(lr){10-11} \cmidrule(lr){12-14} \cmidrule(lr){15-16}
        & & & k=5 & k=10 & k=5 & k=10 & k=5 & k=10 & k=5 & k=10 & TIE $\downarrowRHDSmall$ & LCA $\downarrowRHDSmall$ & Jac $\uparrowRHDSmall$ & AUC $\uparrowRHDSmall$ & AP $\uparrowRHDSmall$ \\
        \midrule
        \clip-S \cite{radford2021learning} & 32.8 & 35.5 & 51.4 & 63.0 & 78.3 & 85.7 & 66.0 & 76.5 & 88.5 & 93.3 & 4.39 & 2.50 & 73.6 & -- & -- \\
        \meru-S \cite{desai2023hyperbolic} & 32.6 & 34.3 & 51.2 & 63.3 & 77.7 & 86.3 & 65.8 & 76.7 & 88.2 & 94.3 & 4.35 & 2.48 & 73.7 & 58.4 & 19.9 \\
        \hycoclip-S \cite{PalSDFGM2024} & 37.3 & \fst{39.2} & 52.0 & 63.5 & 78.5 & 85.9 & 65.2 & 75.9 & 88.7 & 92.8 & 3.96 & 2.33 & 76.2 & 96.8 & 87.7 \\
        \rowcolor{mypink}\methodName-S & \fst{38.8} & 38.4 & \fst{53.2} & \fst{64.7} & \fst{79.9} & \fst{87.5} & \fst{66.9} & \fst{77.8} & \fst{90.4} & \fst{95.6} & \fst{3.76} & \fst{2.26} & \fst{77.7} & \fst{99.3} & \fst{91.2} \\
        $\Delta$ & \inc{+1.5} & \dec{-0.8} & \inc{+1.2} & \inc{+1.2} & \inc{+1.4} & \inc{+1.6} & \inc{+1.7} & \inc{+1.9} & \inc{+0.7} & \inc{+2.8} & \inc{-0.20} & \inc{-0.07} & \inc{+1.5} & \inc{+2.5} & \inc{+3.5} \\
        \midrule
        \clip-B \cite{radford2021learning} & 36.9 & 37.6 & 54.8 & 66.3 & 81.9 & 88.6 & 70.4 & 79.3 & 92.1 & 95.3 & 3.87 & 2.30 & \fst{84.0} & -- & -- \\
        \meru-B \cite{desai2023hyperbolic} & 36.2 & 37.0 & 54.6 & 66.3 & 81.1 & 88.4 & 68.6 & 79.1 & 91.0 & 95.4 & 3.98 & 2.35 & \fst{84.0} & 55.5 & 14.4 \\
        \hycoclip-B \cite{PalSDFGM2024} & 43.1 & 42.6 & 56.7 & 68.0 & 82.8 & 89.3 & 69.5 & 79.7 & 91.6 & 95.9 & 3.40 & 2.11 & 79.9 & 97.2 & 88.5 \\
        \rowcolor{mypink}\methodName-B & \fst{44.0} & \fst{42.8} & \fst{57.9} & \fst{69.2} & \fst{84.4} & \fst{90.7} & \fst{70.8} & \fst{80.9} & \fst{92.9} & \fst{96.2} & \fst{3.36} & \fst{2.10} & 80.2 & \fst{99.4} & \fst{90.6} \\
        $\Delta$ & \inc{+0.9} & \inc{+0.2} & \inc{+1.2} & \inc{+1.2} & \inc{+1.6} & \inc{+1.4} & \inc{+1.3} & \inc{+1.2} & \inc{+1.3} & \inc{+0.3} & \inc{-0.04} & \inc{-0.01} & \inc{+0.3} & \inc{+2.2} & \inc{+2.1} \\
        \midrule
        \clip-L \cite{radford2021learning} & 39.9 & 40.6 & 57.7 & 69.2 & 84.6 & 90.3 & 70.5 & 80.5 & \fst{93.6} & 96.1 & 3.64 & 2.22 & 78.6 & -- & -- \\
        \meru-L \cite{desai2023hyperbolic} & 39.6 & 40.2 & 57.7 & 68.8 & 84.3 & 90.0 & 70.9 & 81.2 & 91.2 & 95.8 & 3.71 & 2.24 & 78.0 & 60.4 & 23.8 \\
        \hycoclip-L \cite{PalSDFGM2024} & 43.9 & 44.4 & 57.5 & 68.6 & 84.2 & 90.1 & 70.7 & 80.2 & 92.3 & 95.9 & 3.32 & 2.09 & 80.5 & 98.0 & 89.5 \\
        \rowcolor{mypink}\methodName-L & \fst{45.6} & \fst{45.1} & \fst{58.6} & \fst{69.7} & \fst{84.9} & \fst{90.6} & \fst{71.7} & \fst{81.7} & 92.8 & \fst{96.8} & \fst{3.15} & \fst{2.03} & \fst{81.6} & \fst{99.5} & \fst{90.3} \\
        $\Delta$ & \inc{+1.7} & \inc{+0.7} & \inc{+1.1} & \inc{+1.1} & \inc{+0.7} & \inc{+0.5} & \inc{+1.0} & \inc{+1.5} & \inc{+0.5} & \inc{+0.9} & \inc{-0.17} & \inc{-0.06} & \inc{+1.1} & \inc{+1.5} & \inc{+0.8} \\
        \bottomrule
    \end{NiceTabular}
    }
\end{table*}

\section{Experiments}\label{sec:experiments}


\mypara{Training Datasets} Our experiments primarily utilize the GRIT dataset \cite{peng2023kosmos}, which comprises $20.5M$ grounded vision-language pairs. We train all models with images and captions, with the exception of \hycoclip, which also incorporates box information. For ablations, we apply \hycoclip's data generation pipeline to create a smaller CC3M dataset~\cite{sharma2018conceptual}.

\mypara{Benchmarks and Metrics} We evaluate our models on a suite of standard downstream tasks including zero-shot image classification and image-text retrieval following previous works~\cite{PalSDFGM2024,desai2023hyperbolic}:
\begin{itemize}
    \item \textbf{Zero-Shot Classification:} We use a set of 16 widely used classification datasets and report top-1 accuracy on ImageNet~\cite{deng2009imagenet}, along with the overall average across all 16 datasets.
    \item \textbf{Image-Text Retrieval:} We report standard metrics, Recall@5 and Recall@10 (R@5, R@10) for both image-to-text and text-to-image retrieval on Flickr30k~\cite{plummer2015flickr30k} and \coco~\cite{lin2014microsoft}.
    \item \textbf{Hierarchical Metrics (WordNet):} Leveraging WordNet~\cite{miller1995wordnet,PalSDFGM2024} to evaluate the hierarchy of ImageNet predictions, we benchmark models using set-based metrics: TIE, LCA, Precision, Recall, and Jaccard Index.
    \item \textbf{Hierarchical Metrics (Ours):} AUC-ROC and AP for our proposed probabilistic entailment protocol.
\end{itemize}


\mypara{Model Architecture} 
We extended the publicly available \hycoclip~\cite{PalSDFGM2024} codebase, employing a dual-encoder \clip architecture~\cite{desai2023hyperbolic,radford2021learning} consistent with established practices. 
The image encoder processes $224{\times}224$ inputs with a $16{\times}16$ patch size, and the text encoder handles a maximum of $77$ tokens. 
This architecture was adapted to hyperbolic space to establish \meru~\cite{desai2023hyperbolic} and \hycoclip~\cite{PalSDFGM2024} as baselines. 
While \meru learns representations for full image-caption pairs, \hycoclip further utilizes phrases and image crops to impose hierarchical structure. 
Our proposed model, \methodName, integrates our adaptive entailment loss ($\mathcal{L}_\text{AdaEnt}$) into the \hycoclip framework. 
For $\mathcal{L}_\text{AdaEnt}$, we empirically set both weighing terms $\gamma_1$ and $\gamma_2$ to $0.1$.
We experimented with Small (S), Base (B), and Large (L) model sizes. 
More details are available in the supplementary material.

\subsection{Quantitative Results}

\Tref{tab:quan} compares \methodName against all baselines trained on the \grit dataset. 
\methodName consistently achieves substantial performance improvements, including gains of up to $2.8\%$ in average zero-shot classification and significant enhancements across all retrieval tasks. 
Furthermore, \methodName shows consistent improvements in hierarchical metrics, specifically the set-based WordNet scores (TIE, LCA, Jaccard), and achieves a superior AUC on our evaluation metrics.
\methodName proves robust across various model scales (Small, Base, Large), consistently achieving top scores in nearly all configurations. 
Notably, increasing model size correlates with more pronounced gains in downstream classification tasks (\eg, a $+1.7\%$ increase on ImageNet for the Large model), while the Average Precision (AP) on HierarCaps decreases. 
This observation suggests a potential trade-off between optimizing for broad downstream performance and fine-grained hierarchical representation, an area reserved for future investigation.

\subsection{Hyperbolic Space Analysis}

\begin{figure}[t]
    \centering
    \vspace{-5pt}
    \begin{subfigure}[t]{0.8\linewidth}
        \includegraphics[width=\linewidth]{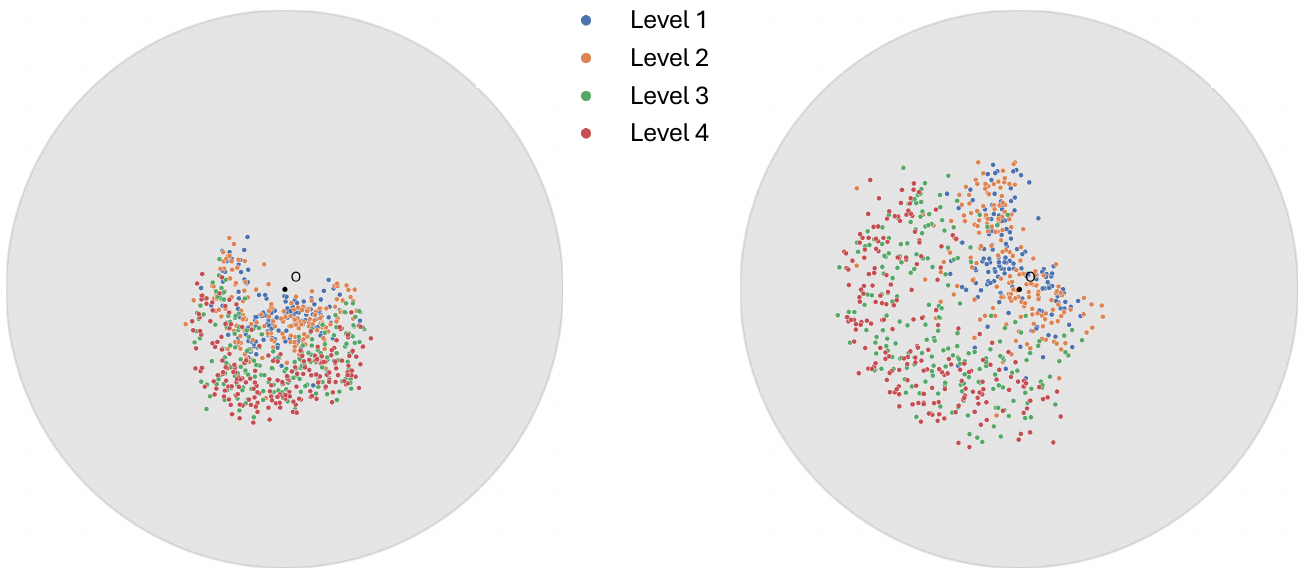}
        \caption{HierarCaps: HoroPCA on different text embedding levels.}
        \label{fig:space_analysis_textlevel}
    \end{subfigure}
    \begin{subfigure}[t]{0.8\linewidth}
        \includegraphics[width=\linewidth]{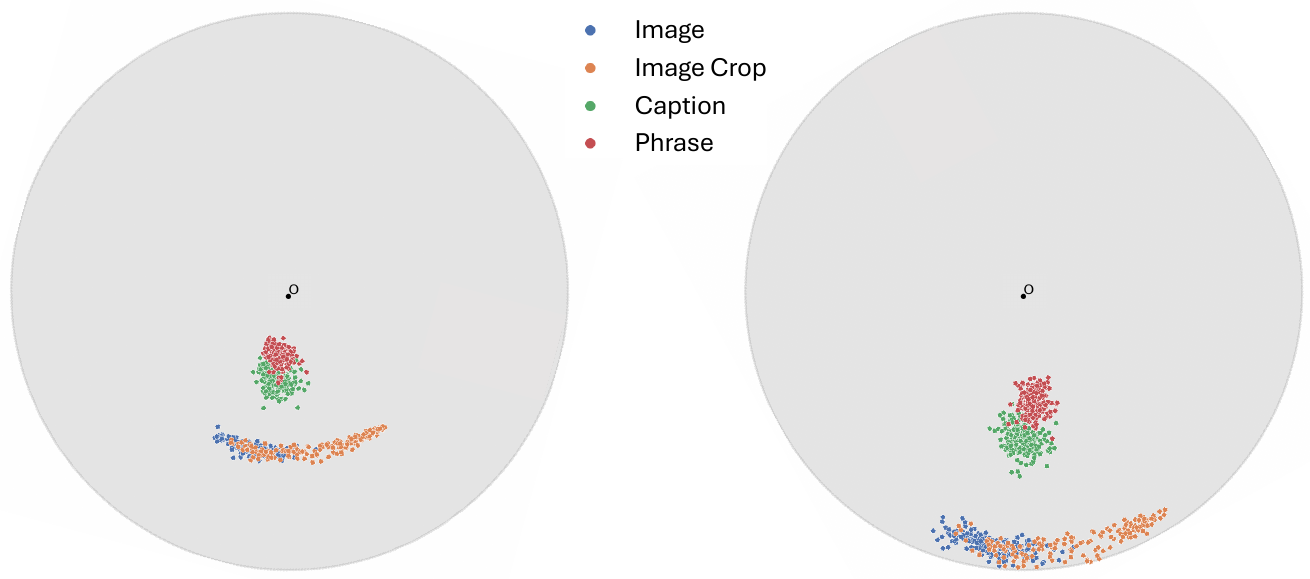}
        \caption{CC3M: HoroPCA on different data components.}
        \label{fig:space_analysis_pca}
    \end{subfigure}
    \begin{subfigure}[t]{0.8\linewidth}
        \includegraphics[width=\linewidth]{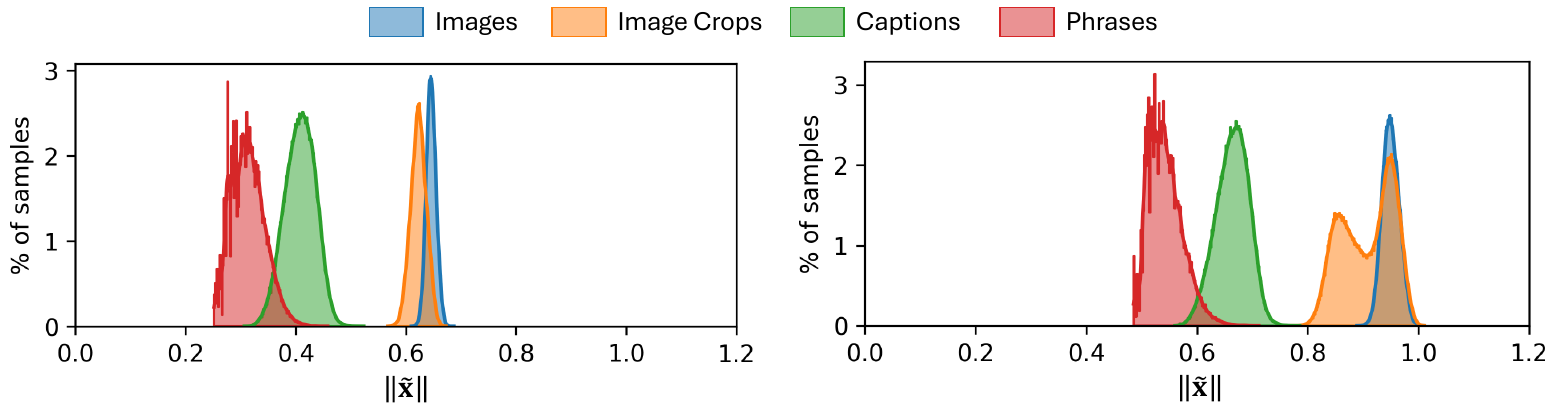}
        \caption{CC3M: Embedding norm distribution.}
        \label{fig:space_analysis_norm}
    \end{subfigure}
    \caption{\textbf{HoroPCA and norm distribution} 
    of \hycoclip (\textit{left}) and \methodName (\textit{right}) on HierarCaps and CC3M.
    \textbf{\methodName increases the variance of embedding norms and reduces the overlap between caption and phrase distributions.}
    }
    \label{fig:space_analysis}
    \vspace{-20pt}
\end{figure}

We next report the impact of the adaptive entailment loss by a comparative analysis of the embedding structures learned by \hycoclip and \methodName. Utilizing HoroPCA~\cite{chami2021horopca}, a specialized hyperbolic dimensionality reduction technique, we project the learned embeddings into 2D Poincaré space and visualize their norm distributions in \Fref{fig:space_analysis}.


\Fref{fig:space_analysis_textlevel} shows HierarCaps text embeddings, revealing \methodName{}'s ability to produce a well-defined hierarchy with clear specificity level separation, unlike \hycoclip{}'s collapsed embeddings near the origin. This highlights the effectiveness of our loss in structuring and stabilizing the hyperbolic embedding space.

Similarly, \Fref{fig:space_analysis_pca}, displaying HoroPCA projections of CC3M components (\textit{images, crops, captions, phrases}), demonstrates \methodName{}'s superior geometric structure by exhibiting clearer decision boundaries between captions and phrases and a broader spatial distribution of embeddings, suggesting a higher-capacity representation space.

Further insights from \Fref{fig:space_analysis_norm}'s norm distributions indicate that \methodName increases embedding norm variance and reduces overlap between caption and phrase distributions. The most significant difference, however, appears in image and crop distributions: while the baseline shows considerable overlap, our model generates a bimodal distribution for crops. One peak aligns with whole images (likely large, salient crops), and a distinct second peak is positioned further away from the origin, showcasing the effectiveness of our adaptive weight $h$.

\begin{table}[t]
    \centering
    \vspace{-0.cm}
    \caption{\textbf{Ablation studies} of proposed components. 
    Our proposed $\mathcal{L}_\text{AdaEnt}$ outperforms standard entailment loss in (a). The adaptive wight is the most efficient when applying on intra-modality pairs in (b). Regularization loss balances the performance of the model on tasks in (d). [Key: \fst{Best} in each group].}
    \label{tab:ablation}
    \footnotesize
    \vspace{-0.2cm}
    \begin{tblr}{width=\linewidth,colspec={@{}X[2,l]|X[1,c]|X[1,c]|X[1,c]|X[1,c]X[1,c]|X[1,c]X[1,c]@{}},colsep=1pt,rowsep=0.7pt}
    \toprule
     \SetCell[r=2]{l}{Model} & \SetCell[r=2]{c}{Module} & Cls. & Ret. & \SetCell[c=2]{c}{WordNet} & & \SetCell[c=2]{c}{PEP} \\
     \hline
     
     & & Acc.~$\uparrowRHDSmall$ & R@5~$\uparrowRHDSmall$ & Err.~$\downarrowRHDSmall$ & Jacc.~$\uparrowRHDSmall$ & AUC~$\uparrowRHDSmall$ & AP~$\uparrowRHDSmall$ \\
     \hline
     \SetCell[c=8]{c} \textit{(a) Adaptive Entailment loss with Different Backbones} \\
     \hline
     \SetCell[r=2]{l}\meru-S & Ent & \fst{20.8} & 52.6 & 5.07 & 59.2 & 73.5 & 45.0  \\
      & AdaEnt & 20.6 & \fst{53.4} & \fst{4.98} & \fst{59.4} & \fst{95.5} & \fst{60.9} \\
     \hline
     \SetCell[r=2]{l}\hycoclip-S & Ent & \fst{22.9} & 54.8 & 4.73 & 62.2 & 94.6 & 76.2 \\
      & AdaEnt & 22.8 & \fst{55.5} & \fst{4.60} & \fst{62.8} & \fst{97.3} & \fst{76.9} \\
     \hline
     \SetCell[r=2]{l}\hycoclip-B & Ent & \fst{25.4} & 59.6 & 4.41 & 64.4 & 94.2 & 76.4 \\
      & AdaEnt & 25.3 & \fst{60.2} & \fst{4.28} & \fst{65.7} & \fst{97.9} & \fst{78.2} \\
     \hline
     \SetCell[r=2]{l}\hycoclip-L & Ent  & 25.8 & 62.1 & 4.37 & \fst{65.5} & 93.6 & 76.0 \\
      & AdaEnt & \fst{26.2} & \fst{63.0} & \fst{4.32} & 65.2 & \fst{97.9} & \fst{79.2} \\
     \hline
     \SetCell[c=8]{c} \textit{(b) Entailment Relationship applied Adaptive Weight $h$}  \\
     \hline
     \SetCell[r=2]{l}\meru-S & None & 20.6 & \fst{53.4} & \fst{4.98} & \fst{59.4}  & \fst{95.5} & \fst{60.9} \\
      & Any & \fst{21.1} & 53.1 & 5.07 & 59.0 & 71.1 & 44.4 \\
     \hline
     \SetCell[r=3]{l}\hycoclip-S & None & 20.9 & 51.3 & 4.87 & 60.4 & 94.9 & 74.3 \\
     & Any & 21.6 & 54.9 & 4.68 & 62.0 & 71.1 & 40.2 \\
     & Intra & \fst{22.8} & \fst{55.5} & \fst{4.60} & \fst{62.8} & \fst{97.3} & \fst{76.9} \\
     \hline
     \SetCell[c=8]{c} \textit{(c) Regularization loss $\mathcal{L}_\text{Reg}$}  \\
     \hline
     \SetCell[r=2]{l}\meru-S & \XSolid & \fst{20.8} & 53.2 & 5.11 & 58.8 & 94.8 & 58.7 \\
     & \Checkmark & 20.6 & \fst{53.4} & \fst{4.98} & \fst{59.4} &  \fst{95.5} & \fst{60.9} \\
     \hline
     \SetCell[r=2]{l}\hycoclip-S & \XSolid & 20.8 & 52.6 & 4.85 & 61.0 & 96.2 & 75.6  \\
     & \Checkmark & \fst{22.8} & \fst{55.5} & \fst{4.60} & \fst{62.8} & \fst{97.3} & \fst{76.9} \\
     \bottomrule
    \end{tblr}
    \vspace{-0.cm}
\end{table}

\subsection{Ablation Studies}


We perform ablation studies on our smaller-scale CC3M dataset, training for $40,000$ iterations ($30M$ data points) to isolate the effect of each component. Unless otherwise specified, we keep all hyperparameters identical to the main experiments and modify only the component under study.




\mypara{Efficacy of Adaptive Entailment Loss} \Tref{tab:ablation}(a) shows that replacing the vanilla entailment loss $\mathcal{L}_\text{Ent}$ (Ent) with $\mathcal{L}_\text{AdaEnt}$ (AdaEnt) consistently improves performance on both \meru and \hycoclip backbones, with particularly strong gains on hierarchical metrics. AdaEnt also boosts downstream retrieval and, for larger models, improves classification performance.

\mypara{Adaptive Weight is Crucial for Intra-Modality Entailment} \Tref{tab:ablation}(b) confirms that applying the adaptive weight $h$ to all pairs (`Any') degrades performance, since it incorrectly weakens the crucial inter-modality constraints. Restricting $h$ to intra-modality pairs only (`Intra') achieves the highest performance across all tasks, validating our choice. 

\mypara{Regularization Loss Stabilizes Performance} \Tref{tab:ablation}(c) shows that removing $\mathcal{L}_\text{Reg}$ lowers performance on both downstream tasks and WordNet-based hierarchical scores. This behavior is expected: without regularization, the embedding space becomes overly sparse, which harms generalization in zero-shot tasks.
\section{Conclusion}

In this work, we tackle key limitations in both the training and evaluation of hierarchical vision-language representations. 
We introduce the Adaptive Entailment Loss ($\mathcal{L}_\text{AdaEnt}$), for learning hierarchical relationships that directly minimizes angular distance, thereby avoiding the constraint violations in prior cone-based losses. 
Concurrently, we propose a new, more reliable evaluation protocol PEP that reformulates hierarchical assessment as a probabilistic entailment classification task.
Our model, \methodName, consistently outperforms previous baselines on most downstream tasks and hierarchical metrics. 
Furthermore, our qualitative analysis of the embedding space, using HoroPCA visualizations, substantiates \methodName{}’s ability to learn a more clearly separated and semantically meaningful hierarchical structure. Our new protocol tackles the critical ambiguities identified in the HierarCaps benchmark. 
By transforming the evaluation from a noisy retrieval task to a probabilistic classification, we leverage standard, robust metrics like AUC-ROC and AP to better quantify a model’s true entailment quality.

\noindent\textbf{Limitations and Future Work:} Our adaptive loss, removes the explicit geometric constraints of prior work, which increases embedding sparsity and necessitates using $\mathcal{L}_\text{Reg}$ regularizer. 
This introduces a new, albeit manageable, trade-off in balancing the different loss components during training. 
Future works involve creating a large-scale, and high-quality benchmark to advance research in hierarchical modeling and retrieval with the new evaluation protocol.

%
%
\bibliographystyle{splncs04}
\bibliography{main}

\clearpage
\appendix
\maketitlesupplementary

\section{Filtering the HierarCaps Benchmark}

As discussed in \Sref{sec:revisit_evaluation}, we use HierarCaps~\cite{alper2024emergent} to evaluate the hierarchy of image-text representations. The benchmark includes 1000 images with 4 captions each of increasing specificity to evaluate whether the embedding captures fine-grained ``is-a'' relationships. The images is from COCO~\cite{lin2014microsoft} while the captions are generated semi-automatically by LLMs and humans. However, HierarCaps contains noisy and ambiguous annotations, which can distort evaluation. We therefore construct a cleaner subset for hierarchical-representation evaluation. Starting from the original benchmark (1000 samples), our filtering yields 522 retained samples.

\mypara{Overview} Our filtering pipeline combines two retrieval models and an MLLM judge to improve robustness:
\begin{enumerate}[label=(\roman*)]
    \item Qwen3-VL-Embedding-8B provides embedding-based retrieval and similarity scoring;
    \item Qwen3-VL-Reranker-8B refines top candidates from the embedding retriever;
    \item Qwen3-VL-8B serves as an MLLM judge to directly assess image-caption alignment.
\end{enumerate}
Throughout filtering, we use the finest-level caption for each sample, as coarse captions are often underspecified and are less reliably assessed by these models.

\mypara{Removing Duplications}
We use Qwen3-VL-Embedding-8B to compute pairwise cosine similarities among images and among captions.
Two samples are treated as duplicates if their image similarity exceeds $0.9$ or text similarity exceeds $0.95$. For each duplicate cluster, we keep one representative sample and remove the rest.  

\mypara{Reducing False Negatives}
Some caption-level negatives in HierarCaps can be false (i.e., a caption may plausibly match multiple images). To identify captions that are likely to be overly specific (and thus safer negatives), we perform image-to-text retrieval using a two-stage process: the embedding model retrieves candidates, and the reranker refines the top-10 results.
For each caption $T$, we define its retrieval degree at $k$ as
\begin{align}
\mathrm{deg}(T)_k \;=\; \bigl|\{\, I \;:\; T \in \mathrm{Top}\text{-}k(\mathrm{Ret}(I)) \,\}\bigr|,
\end{align}
i.e., the number of images $I$ for which $T$ appears in the top-$k$ retrieved captions.
We then select 100 samples whose finest captions have the lowest $\mathrm{deg}(T)_k$ with $k=3$ as negative samples, reducing the probability of including captions that match many images.

\mypara{Reducing False Positives}
To ensure that retained pairs are truly aligned, we apply two complementary checks.

\textit{(1) Bidirectional Retrieval Consistency:}
For each pair $(I_j, T_j)$, we compute:
\begin{itemize}
    \item $\mathrm{rank}_{i2t}(j)$: the rank of $T_j$ among captions retrieved when querying with $I_j$ (image-to-text);
    \item $\mathrm{rank}_{t2i}(j)$: the rank of $I_j$ among images retrieved when querying with $T_j$ (text-to-image).
\end{itemize}
We keep a pair only if it is highly ranked in both directions:
\begin{align}
    \max\bigl(\mathrm{rank}_{i2t}(j),\, \mathrm{rank}_{t2i}(j)\bigr) < 3,
\end{align}
i.e., the ground-truth match appears in the top-2 results for both retrieval directions.

\textit{(2) MLLM Grading:}
To reduce bias from embedding-based retrieval alone, we additionally use Qwen3-VL-8B to score each image-caption pair on a 0-10 scale based on correctness and completeness. The judge prompt is:

\begin{verbatim}
[IMAGE]
You are an expert judge evaluating how well a caption describes 
an image.

Given an image and a caption, evaluate how accurately and completely 
the caption describes the image content.

Caption: [CAPTION]

Provide your judgment with:
1. A rationale explaining the match quality (what matches, what's 
missing, what's incorrect)
2. A match score from 0 to 10:
   - 0: Completely wrong, describes something entirely different
   - 1-3: Poor match, major elements missing or incorrect
   - 4-6: Partial match, captures some elements but misses others
   - 7-8: Good match, captures main elements with minor issues
   - 9-10: Excellent match, accurately describes the image

Answer in JSON format with:
- rationale: Rationale explaining why the caption matches or 
doesn't match the image.
- match_score: Match score between 0 and 10, where 0 is no match and 
10 is perfect match.
\end{verbatim}    

We retain a sample if it satisfies \emph{both} the bidirectional retrieval condition above and an MLLM match score $> 6$. The filtered benchmark will be publicly available upon the acceptance of this work.

\section{Additional PEP Details}
\begin{algorithm}[t]
\caption{PEP Evaluation on Filtered HierarCaps}
\label{alg:pep_eval}
\KwIn{
Embeddings of filtered HierarCaps samples $\mathcal{S}=\{(\mathbf{x}_i,\{\mathbf{y}_i^{\ell}\}_{\ell=1}^{4})\}_{i=1}^{M}$; \\
Embeddings of non-ambiguous negative fine-grained captions $\mathcal{N}=\{\tilde{\mathbf{y}}_k\}_{k=1}^{K}$; }
\KwOut{AUC-ROC and AP.}

\BlankLine
Initialize empty lists of scores and labels: $\mathcal{Z}\leftarrow[\,]$, $\mathcal{T}\leftarrow[\,]$\;

\For{$i \leftarrow 1$ \KwTo $M$}{
    \tcp{Positives: all hierarchy levels for the same image}
    \For{$\ell \leftarrow 1$ \KwTo $4$}{
        $z \leftarrow p_{\mathrm{Ent}}(\mathbf{x}_i, \mathbf{y}_i^{\ell})$\;
        Append $z$ to $\mathcal{Z}$; append label $1$ to $\mathcal{T}$\;
    }

    \tcp{Negatives: fine-grained captions from the non-ambiguous pool}
    \ForEach{$\tilde{\mathbf{y}} \in \mathcal{N}$}{
        \If{$\tilde{\mathbf{y}} = \mathbf{y}_i^{4}$}{
            \textbf{continue}\;
        }
        $z \leftarrow p_{\mathrm{Ent}}(\mathbf{x}_i, \tilde{\mathbf{y}})$\;
        Append $z$ to $\mathcal{Z}$; append label $0$ to $\mathcal{T}$\;
    }
}

Compute \textbf{AUC-ROC} and \textbf{AP} using $(\mathcal{Z}, \mathcal{T})$\;

\Return{AUC-ROC, AP}
\end{algorithm}

PEP is motivated by limitations in the standard HierarCaps evaluation. Kendall’s correlation treats all inversions the same, so it cannot reflect the severity of an ordering error or the strength of entailment margins. PEP instead evaluates entailment directly: it maps the exterior angle $\phi$ to a continuous score $p_\text{Ent}$ and reports AUC-ROC / AP over positive and negative entailment pairs. This protocol is not biased toward ARGENT; it simply rewards geometric consistency under the same entailment geometry. In particular, it favors models that keep positive pairs close to the entailment axis (small $\phi$), rather than merely ``somewhere inside'' a potentially wide cone. ARGENT scores higher because AdaEnt directly minimizes the angle term and avoids reliance on the cone-aperture mechanism (which can require clipping and effectively weaken the constraint), while aperture-based baselines may satisfy the hinge loss with positives near the cone boundary, producing larger angles and thus lower $p_\text{Ent}$.

Some additional implementation clarifications are worth noting. We define the probabilistic entailment score as \Eref{eq:p_ent}. In this regime, we treat the orthogonal (or opposed) embeddings as non-entailed ($p_{Ent}{=}0$). In high-dimensional spaces, orthogonality implies semantic unrelatedness. We also report both AUC-ROC and Average Precision (AP), since they capture complementary behaviors. AUC-ROC mainly measures global separability between positives and negatives, and in our pair construction the labels are highly imbalanced ($\simeq 1: 25$ pos-neg ratio), which can cause the ROC curve to saturate once ``easy'' negatives are rejected correctly. In contrast, AP emphasizes precision on the positive class and is therefore more sensitive to how well the model ranks and concentrates true entailment near the top. Finally, we observe that \meru can score substantially lower under PEP because it is trained only on global image–caption pairs, whereas \hycoclip and our \methodName receive finer-grained hierarchical supervision through additional constructs such as crops and phrases, which better aligns with the entailment pairs evaluated by PEP. The pseudo-code for AUC-ROC, AP is shown in Algorithm~\ref{alg:pep_eval}.

\section{Additional Training Details}
We keep all shared hyper-parameters identical to the original work, with an initial learnable curvature $c{=}1.0$ (clamped within $[0.1, 1.0]$). The learnable scale factor before projecting to the hyperboloid is $1/\sqrt{512}$. The adaptive softmax temperature for the contrastive loss is initialized at $\tau{=}0.07$ and clipped at $0.01$. All learnable scalars are learned in logarithm space. Other variant-specific hyper-parameters are set to the default values reported in their respective papers.

We train all models for $500{,}000$ iterations with a global batch size of $768$, corresponding to approximately $384$ million data points following \cite{PalSDFGM2024}. We use $1$ to $6$ H100 GPUs, depending on the model size. We use the default AdamW optimizer~\cite{loshchilov2019decoupled}, with a weight decay of $0.2$, which is disabled for all learnable scalars. We use a cosine learning rate scheduler with a maximum learning rate of $5 \times 10^{-4}$ and a $4{,}000$-step warm-up.

\section{Quantitative Details}

To complement the summarized results in the main paper (\Tref{tab:quan}), we provide the detailed image classification accuracy across 16 datasets in \Tref{tab:quan_cls}. Specifically, \Tref{tab:quan_cls} reports the performance of our \methodName model across all tested variants. Our method achieves the highest accuracy over 10 datasets when compared against other baselines, demonstrating its strong generalization capabilities. These results validate that the Adaptive Entailment mechanism in \methodName does not negatively impact standard classification tasks.

\begin{table*}[!t]
    \centering
    \caption{\textbf{Image Classification Accuracy on 16 datasets, extension of \Tref{tab:quan}.} \fst{Bold} highlights the best score in each dataset within each model variant.
    [Key: Avg= Average]}
    \label{tab:quan_cls}
    \scalebox{0.7}{
    \setlength{\tabcolsep}{0.07cm} 
    \begin{NiceTabular}{l @{\hspace{12pt}} cccccc @{\hspace{12pt}} cccccc @{\hspace{12pt}} cccc @{\hspace{12pt}} c}
        \toprule
        \multirow{2}{*}{\textbf{\vspace{-1.5cm}Model}} & \multicolumn{6}{c}{\textbf{General}} & \multicolumn{6}{c}{\textbf{Fine-grained}} & \multicolumn{4}{c}{\textbf{Misc}} & \textbf{Avg} \\
        \cmidrule[0.7pt](r{6pt}){2-7} \cmidrule[0.7pt](lr{12pt}){8-13} \cmidrule[0.7pt](lr{12pt}){14-17} \cmidrule[0.7pt](r{6pt}){18-18}
        & \rotatebox{90}{ImageNet} & \rotatebox{90}{CIFAR-10} & \rotatebox{90}{CIFAR-100} & \rotatebox{90}{SUN397} & \rotatebox{90}{Caltech-101} & \rotatebox{90}{STL-10} & \rotatebox{90}{Food-101} & \rotatebox{90}{CUB} & \rotatebox{90}{Cars} & \rotatebox{90}{Aircraft} & \rotatebox{90}{Pets} & \rotatebox{90}{Flowers} & \rotatebox{90}{DTD} & \rotatebox{90}{EuroSAT} & \rotatebox{90}{RESISC45} & \rotatebox{90}{Country211} \\
        \midrule
        \clip-S \cite{radford2021learning} & $32.8$ & $71.6$ & $39.8$ & $45.9$ & $69.3$ & $90.5$ & $44.7$ & $10.2$ & ~$6.4$ & $2.1$ & $38.8$ & $14.6$ & $19.6$ & $37.7$ & $38.8$ & \fst{4.8} & $35.5$ \\ 
        \meru-S \cite{desai2023hyperbolic} & $32.6$ & $72.5$ & $42.0$ & $46.3$ & $69.3$ & $89.9$ & $35.1$ & ~$7.3$ & ~$6.5$ & $2.3$ & $42.3$ & $12.8$ & $18.0$ & $30.2$ & $37.0$ & $4.1$ & $34.3$\\ 
        \hycoclip-S \cite{PalSDFGM2024} & $37.3$ & $83.4$ & $49.2$ & $48.6$ & $71.2$ & $91.6$ & $47.0$ & \fst{12.6} & ~$8.0$ & \fst{2.8} & \fst{47.6} & \fst{19.4} & \fst{22.8} & \fst{42.4} & $39.1$ & $4.7$ & \fst{39.2}\\ 
        \rowcolor{mypink}\methodName-S & \fst{38.8} & \fst{83.6} & \fst{50.5} & \fst{49.2} & \fst{73.7} & \fst{91.9} & \fst{48.0} & $10.3$ & \fst{8.5} & $2.6$ & $43.8$ & $10.9$ & $21.6$ & $37.0$ & \fst{39.4} & $4.7$ & $38.4$\\ 
        \hline
        \clip-B \cite{radford2021learning} & $36.9$ & $71.7$ & $46.4$ & $49.1$ & $74.6$ & $92.6$ & $46.2$ & $10.7$ & ~$8.1$ & $2.0$ & $41.8$ & $18.1$ & $22.1$ & $34.0$ & $41.8$ & $5.2$ & $37.6$\\ 
        \meru-B \cite{desai2023hyperbolic} & $36.2$ & $74.2$ & $45.0$ & $48.5$ & $72.0$ & $91.7$ & $43.2$ & ~$9.9$ & ~$8.5$ & $2.8$ & $44.3$ & $14.5$ & $21.3$ & \fst{35.5} & $39.9$ & $5.1$ & $37.0$\\ 
        \hycoclip-B \cite{PalSDFGM2024} & $43.1$ & $86.8$ & $56.3$ & $54.6$ & \fst{77.8} & $94.1$ & \fst{54.0} & \fst{14.8} & \fst{10.7} & 3.8 & \fst{51.8} & \fst{23.4} & \fst{28.1} & $34.2$ & $42.9$ & $5.6$ & $42.6$\\ 
        \rowcolor{mypink}\methodName-B & \fst{44.0} & \fst{89.0} & \fst{57.8} & \fst{55.1} & 77.5 & \fst{95.3} & $53.3$ & $14.4$ & $10.5$ & \fst{4.1} & 50.6 & 18.0 & 26.0 & 34.6 & \fst{48.9} & \fst{6.1} & \fst{42.8}\\ 
        \hline
        \clip-L \cite{radford2021learning} & $39.9$ & $80.6$ & $51.2$ & $53.4$ & $76.5$ & $93.9$ & $50.4$ & $10.0$ & $10.2$ & $3.2$ & $45.9$ & $17.8$ & $24.7$ & $42.0$ & $44.2$ & $5.8$ & $40.6$\\ 
        \meru-L \cite{desai2023hyperbolic} & $39.6$ & $80.7$ & $51.0$ & $52.7$ & $75.7$ & $94.0$ & $49.4$ & $10.9$ & $10.3$ & $4.4$ & $50.1$ & $19.5$ & $21.7$ & $32.0$ & $46.5$ & $5.5$ & $40.2$\\
        \hycoclip-L \cite{PalSDFGM2024} & $43.9$ & $90.0$ & $61.3$ & $56.2$ & $79.4$ & $94.9$ & $54.7$ & $13.2$ & $11.7$ & \fst{3.8} & \fst{51.4} & \fst{20.3} & \fst{27.5} & \fst{49.2} & $46.3$ & $6.0$ & $44.4$\\ 
        \rowcolor{mypink}\methodName-L & \fst{45.6} & \fst{90.5} & \fst{61.8} & \fst{57.0} & \fst{79.9} & \fst{95.8} & \fst{60.0} & \fst{14.8} & \fst{13.1} & $3.1$ & $51.0$ & $18.6$ & $24.8$ & $48.4$ & \fst{50.2} & \fst{6.3} & \fst{45.1}\\
        \bottomrule
    \end{NiceTabular}
    }
\end{table*}

\section{Qualitative Results}

\begin{figure*}
    \centering
    \includegraphics[height=0.5\linewidth]{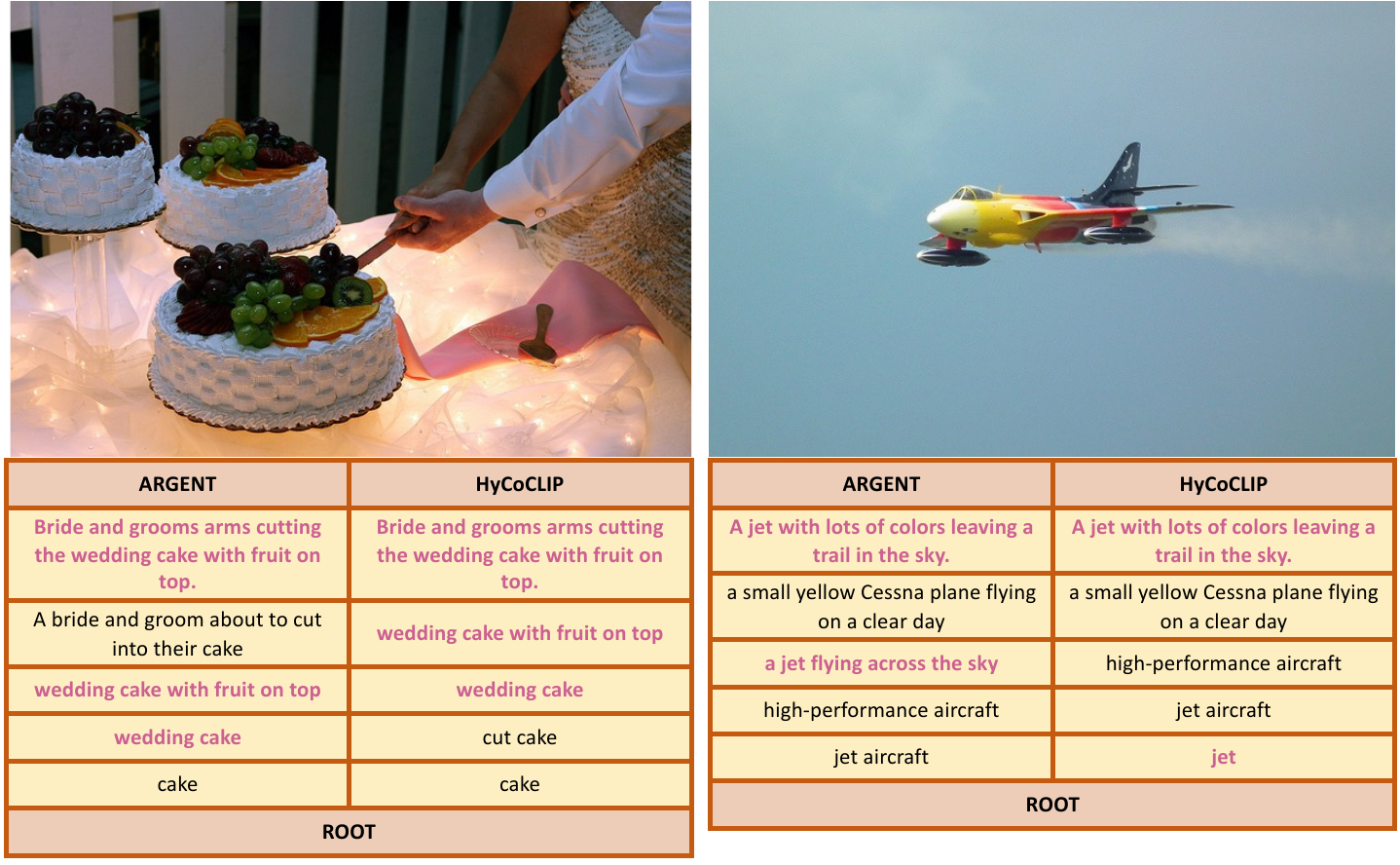}
    \includegraphics[height=0.5\linewidth]{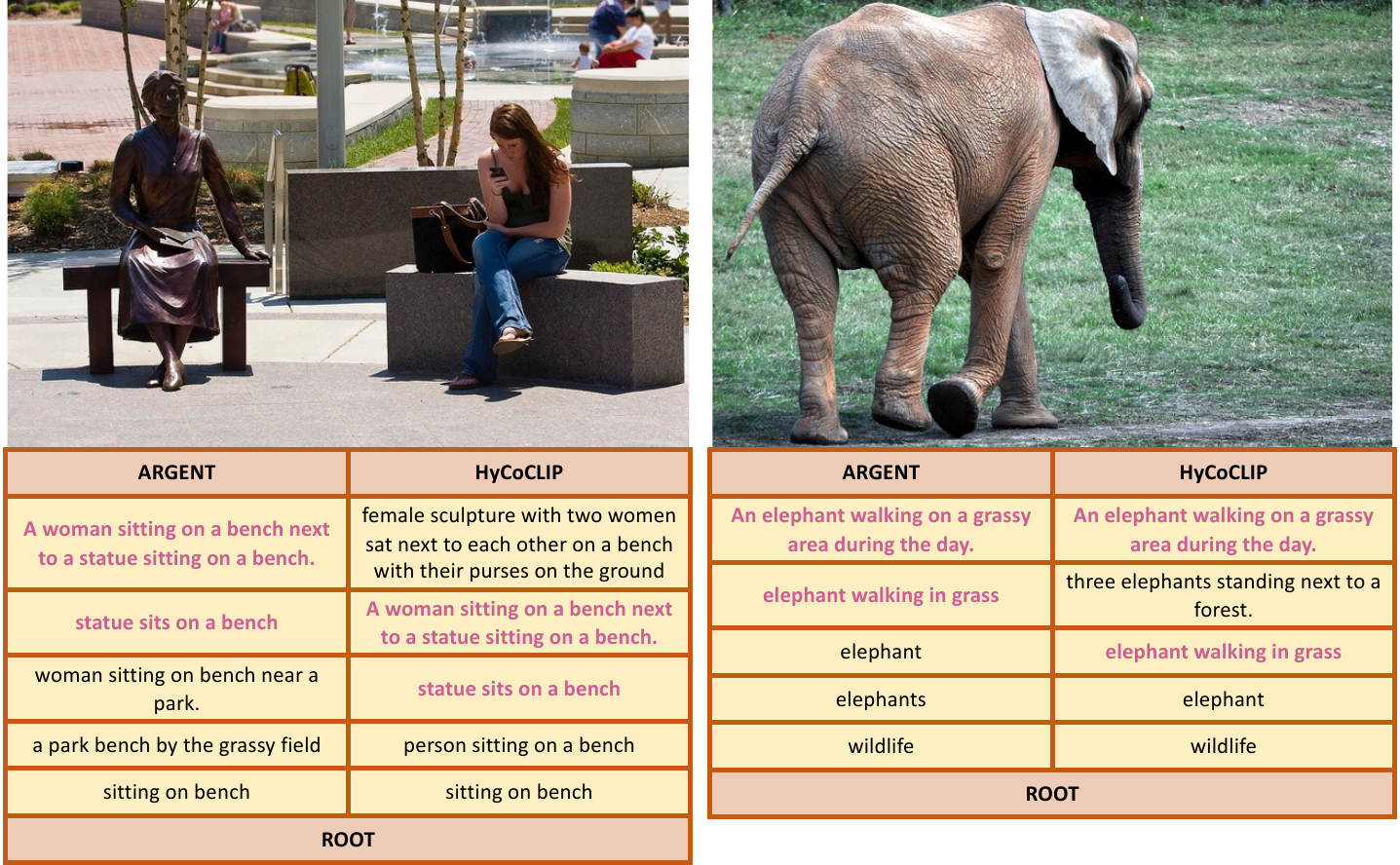}
    \caption{\textbf{Image-to-Text Retrieval Results on HierarCaps dataset.} Our \methodName can retrieve more accurate texts in correct order than the baseline. \textbf{{\color{gt}Purple}} highlights the groundtruth marked in HierarCaps. Groundtruths are in correct order from the most specific to the coarsest detailed captions.}
    \label{fig:qual1}
\end{figure*}

\begin{figure*}
    \centering
    \includegraphics[height=0.5\linewidth]{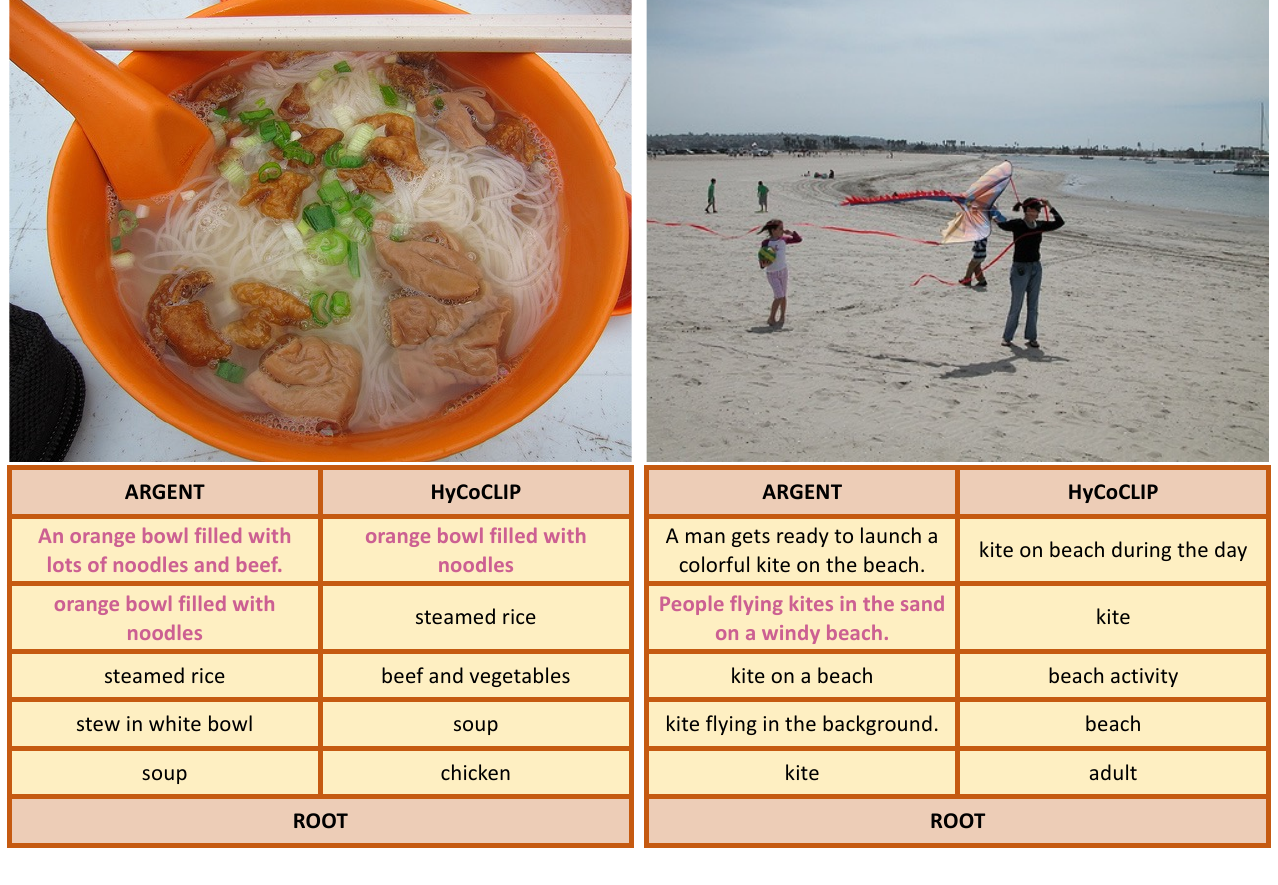}
    \includegraphics[height=0.5\linewidth]{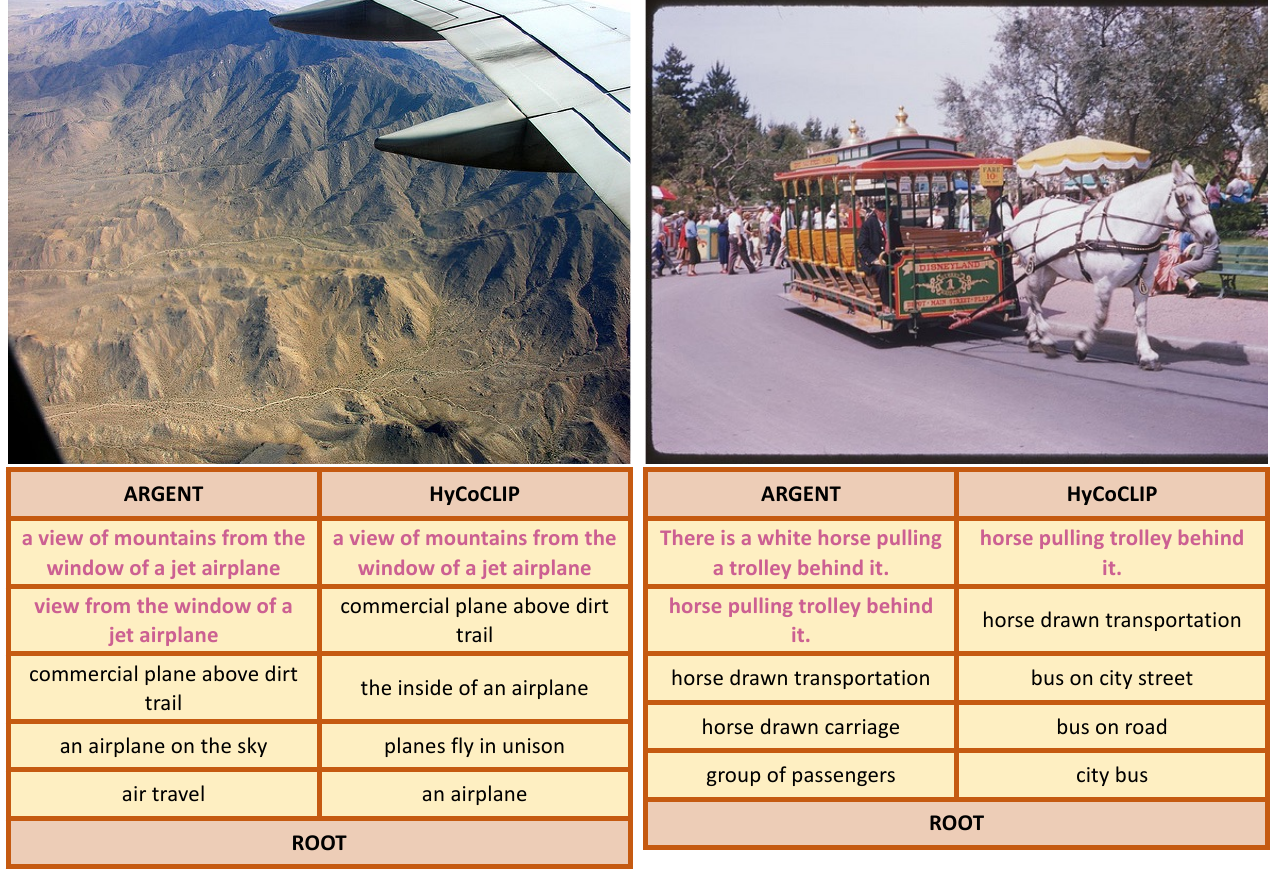}
    \caption{\textbf{Image-to-Text Retrieval Results on HierarCaps dataset. (cont.)} Our \methodName can retrieve more accurate texts in correct order than the baseline. \textbf{{\color{gt}Purple}} highlights the groundtruth marked in HierarCaps. Groundtruths are in correct order from the most specific to the coarsest detail captions.}
    \label{fig:qual2}
\end{figure*}

\Fref{fig:qual1} and \Fref{fig:qual2} present a qualitative comparison between \methodName and \hycoclip on the HierarCaps dataset. We follow the image-to-text retrieval evaluation method detailed in both the original \hycoclip and HierarCaps works. To perform retrieval, the embedding space between the query image and the origin (ROOT) is first divided into 50 equal subspaces. We retrieve the nearest text embedding within each subspace, ignoring any repeated texts. The final retrieved texts are then ordered based on their distance, from the nearest subspace to the furthest. For clarity, we show the top-5 nearest texts to the image in each sample, using the entire HierarCaps text pool as the candidate set.

It is important to note that many of the retrieved texts, though not highlighted as ground truth, are semantically relevant to the images. This observation highlights a limitation of HierarCaps when strictly used as an evaluation benchmark for image-to-text retrieval. Comparing the two methods, our \methodName, with adaptive entailment loss, yields a superior retrieval order. Specifically, we find that more specific captions are ranked nearer to the image, and a higher count of correct texts are retrieved at the top ranks. In contrast, \hycoclip retrieves several inaccurate texts. This likely stems from poor establishment of entailment constraints during \hycoclip's training, which results in noisy entailment relationships between samples and less precise rankings.

\end{document}